\documentclass[11pt,a4paper]{article} 

\usepackage{latexsym}
\usepackage{amssymb} 
\usepackage{amsmath} 
\usepackage{amsthm}
\usepackage{eucal} 
\usepackage{bezier}
\usepackage[matrix,curve,arrow,tips,frame]{xy}

\theoremstyle{plain} \newtheorem{theorem}{Theorem}
\newtheorem{lemma}[theorem]{Lemma}

\newtheorem{proposition}[theorem]{Proposition}
\newtheorem{corollary}[theorem]{Corollary} \newtheorem{cl}{Claim}

\theoremstyle{definition} \newtheorem{definition}[theorem]{Definition}

\theoremstyle{remark}

\newcommand{\N}{\mathbb{N}}

\DeclareMathOperator{\PP}{\mathcal{P}}
\DeclareMathOperator{\LL}{\mathcal{L}}

\newcommand{\NP}{\mathcal{NP}}
\newcommand{\coNP}{\mathrm{co}\mathcal{NP}}

\newcommand{\PSPACE}{\mathcal{PSPACE}}

\newcommand{\X}{\mathcal{X}}
\newcommand{\Y}{\mathcal{Y}}

\newcommand{\Lor}{\bigvee}
\newcommand{\Land}{\bigwedge}

\DeclareMathOperator{\Since}{\mathsf{Since}}

\DeclareMathOperator{\PREV}{\unitlength 1.00pt\begin{picture}(8.00,10.00)\put(4.00,3.00){\circle*{6}}\end{picture}}
\DeclareMathOperator{\PDIA}{\blacklozenge}
\DeclareMathOperator{\PBOX}{\blacksquare}

\newcommand{\TL}{\mathrm{TL}}
\newcommand{\M}{\mathcal{M}} 

\newcommand{\conn}[9]{
\begin{array}{|c|ccc|} \hline
\multicolumn{4}{|c|}{x #1 y}\\ \hline\hline x\diagdown y &0 &1 &\bot\\\hline
 0 &#2 &#3 &#4\\ 1 &#5 &#6 &#7\\ \bot &#8 &#9
&\bot\\ \hline
\end{array}}

\newcommand{\SAC}{\mathrm{SAC}} 
\newcommand{\GNW}{\mathrm{GNW}}
\newcommand{\PS}{\mathrm{PS}} 
\newcommand{\CL}{\mathrm{CL}} 
\newcommand{\Sch}{\mathrm{Sch}}

\newcommand{\lra}{\leftrightarrow}

\newcommand{\true}{1} 
\newcommand{\false}{0}

\newcommand{\A}{\mathfrak{A}}
\newcommand{\B}{\mathfrak{B}}
\newcommand{\E}{\mathcal{E}}

\newcommand{\cea}{{\em cea\/}}
\newcommand{\fM}{\mathfrak{M}}
\newcommand{\RR}{{\mathsf{R}}}
\newcommand{\RS}{{\mathsf{RS}}}
\renewcommand{\S}{{\mathsf{S}}}
\newcommand{\R}{{\mathsf{R}}}
\newcommand{\C}{\complement}
\newcommand{\trzy}{{\text{\boldmath$\mathsf{3}$}}}
\newcommand{\dwa}{{\text{\boldmath$\mathsf{2}$}}}

\newcommand{\CC}{\mathcal{C}}

\newcommand{\first}{\mathrm{first}}

\newcommand{\ce}[1]{[\![#1]\!]}
\newcommand{\ps}[1]{\langle\!\langle#1\rangle\!\rangle_\PS}
\newcommand{\gnw}[1]{\langle\!\langle#1\rangle\!\rangle_\GNW}
\newcommand{\sac}[1]{\langle\!\langle#1\rangle\!\rangle_\SAC}
\renewcommand{\cl}[1]{\langle\!\langle#1\rangle\!\rangle_\CL}
\newcommand{\ttt}[1]{\tau(#1)}
\newcommand{\ttl}[1]{\sigma(#1)}
\newcommand{\ttlr}[1]{\sigma^{\R}(#1)}
\newcommand{\ttr}[1]{\tau^{\RR}(#1)}

\newcommand{\ttrs}[1]{\tau^{\RS}(#1)}
\newcommand{\TT}{{(\TL|\TL)}}
\newcommand{\pr}{\Pr\nolimits}
\newcommand{\defn}{\mathop{\uparrow}}
\newcommand{\TRUE}{\mathit{true}}
\newcommand{\FALSE}{\mathit{false}}

\newcommand{\llnot}{\sim\!}
\title{Embedding Conditional Event Algebras Into  Temporal Calculus
    of Conditionals}
\author{
Jerzy Tyszkiewicz$^{1,2}$\\
Achim Hoffmann$^2$\\
Arthur Ramer$^2$}
\begin{document} 
\begin{titlepage}
\maketitle 
\begin{center}$^1$ Institute of Informatics,\\ Warsaw University,\\
Banacha 2,\\ 02-097 Warszawa,\\ Poland.\\ E-mail {\tt
jty@mimuw.edu.pl}.\\ Supported by the Polish Research Council KBN grant
8 T11C 027 16.\\[5pt] $^2$ School CSE,\\ UNSW,\\ 2052 Sydney,\\
Australia.\\ E-mail {\tt \{jty|achim|ramer\}@cse.unsw.edu.au}.\\
Supported by the Australian Research Council ARC grant A 49800112
(1998--2000).
\end{center}
\thispagestyle{empty}
\end{titlepage}
\begin{titlepage}
\begin{abstract} In this paper we prove that all the existing {\em
conditional event algebras\/} (abbreviated \cea{} in this paper) embed
into the three-valued extension $\TT$ of temporal logic of discrete past
time, which the authors of this paper have proposed in \cite{TT} as a
general model of conditional events.

First of all, we discuss the descriptive incompleteness of the cea's.
In this direction, we show that some important notions, like
independence of conditional events, cannot be properly addressed in
the \cea{} framework, while they can be precisely formulated and
analyzed in the $\TT$ setting.

We also demonstrate that the embeddings allow one to use the native
$\TT$ algorithms for computing probabilities of complex conditional
expressions of the embedded \cea's, and that these algorithms can
outperform those previously known.
\end{abstract}
\end{titlepage}

\begin{titlepage}
\tableofcontents 
\listoffigures
\end{titlepage}

\section{Preliminaries and statement of the problem}

\subsection{The problem of conditional objects}

Probabilistic reasoning \cite{p88} is the basis of Bayesian methods of
expert system inferences, of knowledge discovery in databases, and in
several other domains of computer, information, and decision
sciences.  The model of conditioning and conditional objects we discuss
serves equally to reason about probabilities over a finite domain $X$,
or probabilistic propositional logic with a finite set of atomic
formulae.

Computing of conditional probabilities of the form
$\Pr(X|Y_1,\dots,Y_n)$ and, by extension of conditional beliefs, is well
understood.  Attempts of defining first the {\em conditional objects\/}
of the basic form $X|Y$, and then defining $\Pr(X|Y)$ as $\Pr((X|Y))$
were proposed, without much success, by some of the founders of
probability \cite{b57,d72}.  They were taken up systematically only
about 1980 \cite{a86,l76,es94,g91,gn96}.  The development was slow, both
because of logical difficulties --- interpretation of conditionals, and
even more because the computational model is difficult to construct.
(While $a|b$ appears to stand for a sentence `if $b$ then $a$' and the
probability is $\Pr(a|b)=\Pr(a\land b)/\Pr(b)$, there is no obvious
calculation for $\Pr(a|(b|c))$, nor intuitive meaning for $a|(b|c),$
$(a|b)\land(c|d),$ and the like.)

The idea of defining conditional objects was entertained by some
founders of modern probability \cite{b57,d72}, but generally abandoned
since introduction of the measure-theoretic model.  It was revived
mostly by philosophers in 1970's \cite{a86,v77} with a view towards
artificial intelligence reasoning.  Formal computational models came in
the late 1980's \cite{c87,gnw91} with only one, based on formal
fractions and three-valued indicator functions, used for few actual
calculations of conditionals and their probabilities.  That model may
give results whose values are open to questions \cite{c94}.

The authors of this paper have developed in the companion paper
\cite{TT} a temporal calculus $\TT$ of conditionals, based on
the early ideas of de Finetti \cite{d72}.

In the present paper we show that all the major previously existing
systems of conditionals, the so called {\em conditional event algebras}
(see \cite{kniga} or Sect.\ \ref{cea} in the current paper), embed
isomorphically into $\TT.$ Looking at them as fragments of $\TT,$ we
demonstrate their insufficient expressive power and other defects in
their construction.

They attempt to provide certain kind of a model of the logic of
conditional expressions, built up from simple conditionals of the form
$(a|b)$ with the connectives: conjunction, disjunction and
complementation.

However, the semantical objects assigned to the expressions are not
required to be of probabilistic nature, so they fail to provide
methods to verify the chosen structure experimentally.

The structure of conditionals is not determined functorially by the
space of nonconditional  events. 

Moreover, very restricted setting of \cea's does not allow one to
address many important questions, like stochastic independence of
complex conditionals. In the $\TT$ setting independence can be precisely
defined and analyzed, unlike in the \cea\ formalism, (cf.\ Theorem
\ref{PS-indep} below).

Finally, we use the  algorithms for calculating probabilities
in $\TT,$ which stem from the highly developed algorithms for
calculating limiting probabilities in Markov chains, and apply them to
the embedded \cea's. It appears that these algorithms clearly
outperform the previously known ones for the important {\em product
space} \cea{} \cite{g94}.

Consequently, we believe that our $\TT$ can be used as a single
alternative to each of the major \cea's considered in the literature so
far, superior to each of them,  in the sense of expressive power,
clarity of logical and semantical structure, and, last but not least,
availability of efficient algorithms.

\section{The tools}
\subsection{$\TT$, Moore machines and Markov chains}

We describe briefly the construction of temporal conditionals,
presented in detail in \cite{TT}.

Let $\E=\{a,b,c,d,\dots\}$ be a finite set of basic events, and let
$\Sigma$ be the Boolean algebra generated by $\E,$ and $\Omega$ the set
of atoms of $\Sigma.$ Consequently, $\Sigma$ is isomorphic to the
powerset of $\Omega,$ and $\Omega$ is isomorphic to the powerset of
$\E.$ Any element of $\Sigma$ will be considered as an event, and, in
particular, $\E\subseteq\Sigma.$

The union, intersection and complementation in $\Sigma$ are denoted by
$a\cup b,\ a\cap b$ and $a^\C,$ respectively. The least and greatest
elements of $\Sigma$ are denoted $\varnothing$ and $\Omega,$
respectively. However, sometimes we use a more compact notation,
replacing $\cap$ by a juxtaposition. When we turn to logic, it is
customary to use yet another notation: $a\lor b,\ a\land b$ and $\lnot
a,$ respectively. In this situation $\Omega$ appears as $\TRUE$ and
$\varnothing$ as $\FALSE,$ but $1$ and $0,$ respectively, are
incidentally used, as well.

$\trzy =\{0,1,\bot\}$ is the set of truth values, interpreted as {\em
true}, {\em false\/} and {\em undefined}, respectively.  The subset of
$\trzy $ consisting of $0$ and $1$ will be denoted $\dwa.$

\bigskip

Let us first define {\em temporal logic of linear discrete past time},
called $\TL.$ 

The formulas are built up from the set $\E$ (the same set of basic
events as before), interpreted as propositional variables here, and
are closed under the following formula formation rules:

\begin{enumerate}
\item Every $a\in \E$ is a formula of temporal logic.
\item If $\varphi,\psi\in \TL,$ then their boolean combinations
$\varphi\lor \psi$ $\lnot\varphi$ are in $\TL.$ The other Boolean
connectives: $\land,\to,\lra,\dots$ can be defined in terms of $\lnot$
and $\lor,$ as usual.

\item If $\varphi,\psi\in \TL,$ then their past tense temporal
combinations $\PREV\varphi$ and $\varphi\Since\psi$ are in $\TL,$
where $\PREV\varphi$ is spelled ``previously $\varphi.$''
\end{enumerate}

A model of temporal logic is a sequence $\M=s_0,s_1,\dots,s_n$ of
states, each state being a function from $\E$ (the same set of basic
events as before) to the boolean values $\{0,1\}.$ Note that a state
can be therefore understood as an atomic event from $\Omega,$ and $\M$
can be thought of as a word from $\Omega^+.$ The states of $\M$ are
ordered by $\leq,$ and $s+1$ denotes the successor state of $s.$ We
adopt the convention that, unless explicitly indicated otherwise, a
model is always of length $n+1,$ and thus $n$ is always the last state
of a model.

For every state $s$ of $\M$ we define inductively what it means that a
formula $\varphi\in\TL$ is satisfied in the state $s$ of $\M,$
symbolically $\M,s\models\varphi.$

\begin{enumerate}
\item $\M,s\models a$ iff $s(a)=1$
\item
\begin{align*}
\M,s\models\lnot\varphi&:\iff\ \M,s\not\models\varphi,\\
\M,s\models\varphi\lor\psi&:\iff
\M,s\models\varphi\ \text{or}\ \M,s\models\psi.
\end{align*}

\item
\begin{align*}
\M,s\models\PREV\varphi&:\iff s>0\ \text{and}\ \M,s-1\models\varphi;\\
\M,s\models\varphi\Since\psi&:\iff(\exists t\leq s)(\M,t\models\psi\
\text{and}\ (\forall t< w\leq s)M,w\models\varphi).
\end{align*}
\end{enumerate}

The syntactic abbreviations $\PBOX\varphi$ and $\PDIA\varphi$ are of
common use in $\TL.$ They are defined by $\PDIA\varphi\equiv\TRUE
\Since\varphi$ and $\PBOX\varphi\equiv\lnot\PDIA\lnot\varphi.$ The
first of them is spelled ``once $\varphi$'' and the latter ``always in
the past $\varphi$''.
 
Their semantics is then equivalent to
\begin{align*}
\M,s\models\PBOX\varphi&\iff(\forall t\leq s)\M,t\models\varphi;\\
\M,s\models\PDIA\varphi&\iff(\exists t\leq s) \M,t\models\varphi.
\end{align*}

\begin{theorem}[see \cite{Emerson}]\label{Emerson}
The set of valid $\TL$ formulas is complete in $\PSPACE.$ The set of
valid $\TL$ formulas with $\PBOX$ and $\PDIA$ as the only temporal
connectives is complete in $\coNP.$
\end{theorem}

$\TT$ is the logic of formulas of the form $(\varphi|\psi),$ where
$\varphi,\psi\in\TL.$ $\TT$ is a $\trzy$-valued extension of $\TL,$
and $(\varphi|\psi)$ is 

\begin{itemize}
\item[$1).$] true in $\M,n$ iff $\M,n\models\varphi\land\psi.$
\item[$0).$] false in $\M,n$ iff $\M,n\models\lnot\varphi\land\psi.$
\item[$\bot).$] undefined in $\M,n$ iff $\M,n\models\lnot\psi.$
\end{itemize}

\bigskip

A {\em $\trzy$-valued Moore machine $\A$} is a five-tuple
$\A=(Q,\Omega,\delta,h,q_0),$ where where $Q$ is its set of states,
$\Omega$ (the same set of atomic events as before) is the input
alphabet, $q_0\in Q$ is the initial state and $\delta: Q
\times\Omega\to Q$ is the transition function, and $h$ is the output
function $Q\to\trzy.$

Formally, to describe the computation of $\A$ we extend $\delta$ to a
function $\hat{\delta}: Q \times\Omega^+\to Q$ in the
following way: 

\[\hat{\delta}(q,w)=\begin{cases}
\delta(q,w)&\text{if $|w|=1$}\\
\delta(\hat{\delta}(q,v),\omega)&\text{if $w=v\omega.$}
                    \end{cases}
\]

$\A$ computes a function $f_\A:\Omega^+\to \trzy^+$ defined by

\[
f_\A(\omega_1\omega_2\dots\omega_n)
=h(\hat{\delta}(q_0,\omega_1))h(\hat{\delta}(q_0,\omega_1\omega_2))\dots
h(\hat{\delta}(q_0,\omega_1\omega_2\dots\omega_n))
\]

(note that $|f_\A(\omega_1\omega_2\dots\omega_n)|=n,$ as desired)

We picture $\A$ as a labeled directed graph, whose vertices are
elements of $Q,$ labeled by their values under $h,$ the function
$\delta$ is represented by directed edges labeled by elements of
$\Omega$: the edge labeled by $\omega\in\Omega$ from $q\in Q$ leads
to $\delta(q,\omega).$ The initial state is typically indicated by an
unlabeled edge ``from nowhere'' to this state.

As the letters of the input word $w\in\Omega^+$ come in one after
another, we walk in the graph, always choosing the edge labeled by
the letter we receive. At each step it reports to the outside world
the value $h(q)$ of the state $q$ in which it is at the moment.

Drawing Moore machines, we almost always make certain graphical
simplification: we merge all the transitions joining the same pair of
states into a single transition, labeled by the union (evaluated in
$\Sigma$) of all the labels. Sometimes we go even farther and drop the
label altogether from one transition, which means that all the
remaining input letters follow this transition.

It is known for deterministic finite automata \cite{HU}, and extends
easily to Moore machines with the same proof, that for any such device
there is a unique (up to isomorphism) {\em minimal\/} (with respect to
the number of states) device of the same kind, which accepts the same
language (computes the same function, respectively).  Moreover, this
minimal device can be obtained from any such device as a quotient
automaton/machine, i.e., by dividing the state space by some equivalence
relation. For details, including a very efficient algorithm to perform
minimization, see \cite{HU}.

\begin{definition}
A Moore machine $\A$ is called {\em counter-free\/} if there is no
word $w\in\Omega^+$ and no states $q_1,q_2,\dots,q_s,\ s>1,$ such that
$\hat\delta(q_1,w)=q_2,\dots,\hat\delta(q_{s-1},w)=
q_s,\hat\delta(q_s,w)=q_1.$
\end{definition}

Sometimes we use an extension of Moore machines---{\em Moore machines
with $\epsilon$-moves.} In a deterministic finite automaton an
$\epsilon$-move is a transition between two states done without
intervention of any letter from the input. By necessity, to maintain
the deterministic character of the automaton, an $\epsilon$-move must
not be combined with any other transitions starting from the same
state. For Moore machines, we adopt the convention that after
performing an $\epsilon$-move, no symbol is appended to the output.

E.g., for the Moore machine with $\epsilon$-moves $\A$ below

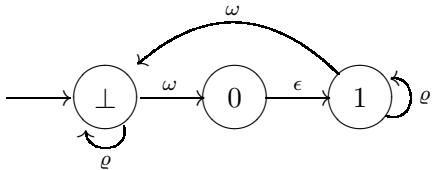
\[\UseTips
\xymatrix
{
{}\ar[r]&
*+++[o][F]{\bot}
\ar@(dr,dl)[]^{\varrho}
\ar[r]^\omega&
*+++[o][F]{0}
\ar[r]^{\epsilon}&
*+++[o][F]{1}
\ar@(dr,ur)[]_{\varrho}
\ar@/_1cm/[ll]_{\omega}
}
\]

we have $f_{\A}(\omega\omega\omega)=0\bot0$ and
$f_{\A}(\omega\varrho\omega)=01\bot.$

It is known that any function computable by a Moore machine with
$\epsilon$-moves can be computed by a Moore machine without
$\epsilon$-moves. Thus using $\epsilon$-moves we do not achieve
greater generality. However, some transformations of the machines can
be very conveniently represented by introducing $\epsilon$-moves.

\bigskip

For us, Markov chains are a synonym of {\em Markov chains with
stationary transitions and finite state space.}

Formally, given a finite set $I$ of {\em states} and a fixed function
$p:I\times I\to[0,1]$ satisfying \( (\forall i\in I)\qquad\sum_{j\in
I}p(i,j)=1, \) the {\em Markov chain\/} with state space $I$ and
transitions $p$ is a sequence $\X=X_0,X_1,\dots$ of random variables
$X_n:W\to I$, such that

\begin{equation}\label{M1}\Pr(X_{n+1}=j|X_n=i)=p(i,j).\end{equation}

The standard result of probability theory is that there exists a
probability triple $(W,\fM,\Pr)$ and a sequence $\X$ such that
\eqref{M1} is satisfied. $W$ is indeed the space of infinite sequences
of ordered pairs of elements from $I,$ and $\Pr$ is a certain product
measure on this set.

One can arrange the values $p(i,j)$ in a matrix $\Pi=(p(i,j);i,j\in
I).$ Of course, $p(i,j)\geq 0$ and $\sum_{j\in I}p(i,j)=1$ for every
$i.$ Every real square matrix $\Pi$ satisfying these conditions is
called {\em stochastic.}

The initial distribution of $\X$ is that of $X_0,$ which can be
conveniently represented by a vector $\Xi_0=(p(i); i\in I).$
Its choice  is independent from the function $p(i,j).$

It follows by a simple calculation that

\begin{equation}\label{MP}
\Pr(X_{n+1}=j|X_n=i_n,X_{n-1}=i_{n-1},\dots,X_1=i_1)
=\Pr(X_{n+1}=j|X_n=i), 
\end{equation}
which is called the {\em Markov property.}

For our purposes, it is convenient to imagine the Markov chain $\X$ in
another, equivalent form: Let $K_I$ be the complete directed graph on
the vertex set $I.$ First we randomly choose the starting vertex in
$I,$ according to the initial distribution.  Next, we start walking in
$K_I;$ at each step, if we are in the vertex $i,$ we choose the edge
$(i,j)$ to follow with probability $p(i,j).$ If we define
$X_n=(\text{the vertex in which we are after $n$ steps}),$ then $X_n$
is indeed the same $X_n$ as in \eqref{M1}.

So we will be able to {\em draw\/} Markov chains.  Doing so, we will
often omit edges $(i,j)$ with $p(i,j)=0.$

\subsection{Conditional objects and conditional events}

Let for $(\varphi|\psi)\in\TT$ the function
$c=c_{(\varphi|\psi)}:\Omega^+\to\trzy$ be defined by

\[c(w)=\begin{cases}1&\text{if $(\varphi|\psi)$ is true in $(w,n)$}\\
                    0&\text{if $(\varphi|\psi)$ is false in $(w,n)$}\\
                    \bot&\text{if $(\varphi|\psi)$ is undefined in $(w,n)$} 
       \end{cases}
\]

$\CC$ is the set of all functions $c:\Omega^+\to\trzy$ definable in
$\TT,$ and $\CC_+$ is the set of all functions
$c_+:\Omega^+\to\trzy^+$ computable by counter-free $\trzy$-valued
Moore machines.

$\CC$ and $\CC_+$ are isomorphic under the mapping $\CC_+\ni c_+\mapsto
c\in\CC$ defined by 

\[c(w)=\text{last-letter-of}(c_+(w)).\]

The sets $\CC$ and $\CC_+$ are regarded as two representations of {\em
conditional objects}. We have yet another representation, denoted
$\CC_\infty$: it consists of functions $\Omega^\infty\to\trzy^\infty,$
and $c_\infty,$ the third representation of the same conditional, is an
infinite sequence of values of $c$ on all finite nonempty prefixes of
$w.$

We will be using the name {\em conditional events\/} to refer to
conditionals considered with a probability space in the background.
\bigskip

\begin{definition}[Conditional event \cite{TT}] Let $c\in \CC$ be a
conditional object over $\Omega$ and let $\A$ be any counter-free
Moore machine with output function $h,$ computing $c_+.$ Suppose
$\Omega$ is endowed with a probability space structure
$(\Omega,\fM,\Pr).$ The conditional object $c$ becomes then a sequence
$\ce{c}=\ce{c}_1,\ce{c}_2,\dots$ of random variables
$\ce{c}_n:\Omega^\infty\to\trzy,$ defined by the formula

\begin{equation}\label{e1}
\ce{c}_n(w)=c(\text{prefix-of-length-$n$-of}(w)),
\end{equation}

where $\Omega^\infty$ is considered with the product probability structure.

We call $\ce{c}$ the {\em conditional event\/} associated with $c.$

Moreover, in presence of probability space structure $\A$ becomes a
Markov chain $\X(\A)$ (by replacing labels of th e transitions by
their probabilities under $\Pr$ in the diagram of $\A$), and then
$\ce{c}=h(\X),$ where $h$ is the output function of $\A.$
\end{definition}

In particular, $\Pr(\ce{c}_n=\true)$ is the probability that at time
$n$ the conditional object $c$ is true, $\Pr(\ce{c}_n=\false)$ is the
probability that at time $n$ the conditional object $c$ is false, and
$\Pr(\ce{c}=\bot)$ is the probability that at time $n$ the conditional
object $c$ is undefined.

\begin{definition}[asymptotic probability \cite{TT}]
We define the {\em asymptotic probability at time $n$} of a
conditional event $c\in\CC$ by the formula

\begin{equation}\label{e2}
\pr_n(c)=\dfrac{\Pr(\ce{c}_n=1)}
{\Pr(\ce{c}_n=0\ \text{or}\ 1)}.
\end{equation}

If the denominator is $0,$ $\pr_n(c)$ is undefined.

The {\em asymptotic probability\/} of $c$ is

\begin{equation}\label{e3.5}
\Pr(c)=\lim_{n\to\infty}\pr_n(c),
\end{equation}

provided that $\pr_n(c)$ is defined for all sufficiently large $n$ and
the limit exists.

If $\varphi\in\TL$ then we write $\Pr(\varphi)$ for $\Pr((\varphi|\TRUE)).$ 
\end{definition}

\begin{theorem}[Bayes' Formula \cite{TT}]\label{Bayes}

Let $(\varphi|\psi)$ be a conditional object over $\Omega$ endowed
with a probability space structure $(\Omega,\fM,\Pr),$ and let $\A$ be
any counter-free Moore machine with output function $h,$ computing
$(\varphi|\psi).$
\begin{enumerate}
\item For every state $i$ of $\X(\A)$ the probability
$\lim_{n\to\infty}\Pr(X_n=i)$ exists.
\item For every  $\star\in\trzy$  the probability
$\lim_{n\to\infty}\Pr(\ce{(\varphi|\psi)}_n=\star)$ exists.
\item The Bayes' Formula 
\end{enumerate}

\[\Pr((\varphi|\psi))=\frac{\Pr(\varphi\land\psi)}{\Pr(\psi)}\]

holds whenever the right-hand-side above is well-defined, i.e., 
$\Pr(\psi)>0.$
\end{theorem}

\section{Connectives of conditionals}

If we wish to extend the classical two-valued conjunction to
conditionals, we are faced with the problem of {\em
synchronization}. Indeed, what is easy in $\dwa$-valued world becomes
messy in $\trzy $-valued world. The problem is that the conditionals
need not become defined synchronously. For the classical conjunction
this problem does not exist, because both arguments are always
defined. Now we have to resolve the question how to define the
conjunction, when some of the arguments are undefined.

\subsection{Present tense connectives}\label{teraz}

Let us recall that present tense connectives are those, whose definition
in $\TT$ does not use temporal connectives, and therefore depends
on the present, only. Equivalently, an $n$-ary present tense connective is
completely characterized by a function $\trzy^n\to\trzy .$ 

Here are several possible choices for the conjunction, which is always
defined as a pointwise application of the following $\trzy$ valued
functions. Above we display the notation for the corresponding kind of
conjunction.

\[\label{toto}
\begin{array}{ccc}\conn{\land_{\SAC}}00001101& 
\conn{\land_{\GNW}}00001\bot0\bot&
\conn{\land_{\Sch}}00\bot01\bot\bot\bot\\
&&\\
&{\begin{array}{|c|c|}\hline 
\multicolumn{2}{|c|}{\lnot x} \\\hline\hline x&\lnot x\\\hline 0&1\\
1&0\\\bot&\bot\\\hline\end{array}}&\\ &&\\\conn{\lor_{\SAC}}01011101&
\conn{\lor_{\GNW}}01\bot111\bot1&
\conn{\lor_{\Sch}}01\bot11\bot\bot\bot.
\end{array}\]

They can be equivalently described by syntactical manipulations in $\TT.$
The reduction rules are as follows:

\begin{equation}\label{TLTL}
\begin{split}
(a|b)\land_\SAC(c|d)&=(abcd\lor abd^\C\lor cdb^\C|b\lor d)\\
(a|b)\land_\GNW(c|d)&=(abcd|a^\C d\lor c^\C d\lor abcd)\\
(a|b)\land_\Sch(c|d)&=(abcd|bd)\\
\llnot_0 (a|b)&=(a^\C|b)\\
(a|b)\lor_\SAC(c|d)&=(ab\lor cd|b\lor d)\\
(a|b)\lor_\GNW(c|d)&=(ab\lor cd|ab\lor cd \lor bd)\\
(a|b)\lor_\Sch(c|d)&=(ab\lor cd|bd).
\end{split}
\end{equation}

The first is based on the principle ``if any of the arguments becomes
defined, act!''.  A good example would be a quotation from \cite{c97}:

\begin{quote}\sl ``One of the most dramatic examples of the unrecognized
use of compound conditioning was the first military strategy of our
nation.  As the Colonialists waited for the British to attack, the
signal was `One if by land and two if by sea'.  This is the conjunction
of two conditionals with uncertainty!''\footnote{NB, if the British had
decided to attack from both directions, but not simultaneously, we would
have probably discovered temporal conditionals much earlier.}
\end{quote}

Of course, if the above was understood as a conjunction of two
conditionals, the situation was crying for the use of $\land_\SAC,$
whose definition has been proposed independently by Schay, Adams and
Calabrese (the author of the quotation).

The conjunction $\land_{\GNW}$ represents a moderate approach, which
in case of an apparent evidence for $\false$ reports $\false,$ but
otherwise it prefers to report unknown in a case of any doubt. Note
that this conjunction is essentially the same as {\em lazy
evaluation},  known from programming languages. 

Finally, the conjunction $\land_\Sch$ is least defined, and acts
(classically) only if both arguments become defined. It corresponds to
the {\em strict evaluation}.

We have given an example for the use of $\land_\SAC.$ The uses of
$\land_\GNW$ and $\land_\Sch$ can be found in any computer program
executed in parallel, which uses either lazy or strict evaluation of
its logical conditions. And indeed both of them happily coexist in
many programming languages, in that one of them is the standard
choice, the programmer can however explicitly override the default and
choose the other evaluation strategy.

This seems to suggest that neither of the three choices discussed in
this paragraph is {\em the\/} conjunction of conditionals. There are
indeed many possible choices, and all of them have their own merits.
And indeed already the original system of Schay consisted of five
operations:  $\llnot_0,\land_\SAC,\lor_\SAC,\land_\Sch$ and $\lor_\Sch.$
Moreover, he was aware that these operations still do not make the
algebra functionally complete (even in the narrowed sense, restricted to
defining only operations which are undefined for all undefined
arguments). And in order to remedy this he suggested to use one of
several additional operators, one of them being $\land_\GNW!$ So for him
all those operations could coexist in one system.

\paragraph{Present tense re-conditioning}

Calabrese \cite{C90} and Goodman, Nguyen and Walker \cite{gn96} proposed
their own extensions of the conditioning operator to $\trzy,$ hence
making it available for re-conditioning in $\SAC$ and $\GNW$,
respectively.  The definitions are

\[ \conn{|_\SAC}\bot\bot\bot01\bot01\ \ \
\conn{|_\GNW}\bot\bot\bot01\bot0\bot\]

\subsection{Past tense connectives} 

Now we consider connectives, whose definitions refer to the strict
past of their arguments. We continue to consider conjunction, which we
use as a kind of model example.

\paragraph{Examples of past tense connectives}

The following are connectives very close to the conjunction and
disjunction of the {\em product space\/} \cea\/ introduced in
\cite{g94}\footnote{See the discussion on embedding \cea's in our model
below.}, defined by the rule: the conjunction is defined and true iff both of
its arguments have been defined, and moreover the historically {\em first\/}
values of its two arguments have been both $1.$ Otherwise it is defined and
false.  Disjunction is defined similarly.  They use the ``Russian roulette''
approach to repeating experiments.

In the language of $\TT$ $(a|b)\land_\PS (c|d)$ can  be expressed by

\begin{equation}\label{land_PS}
\left(\left.
\begin{array}{c}
\PDIA(a\land b\land
(\lnot\PREV\PDIA b))\\
\land\\
\PDIA(c\land d\land
(\lnot\PREV\PDIA d))
\end{array}
\right|
\TRUE
\right),
\end{equation}

and the definition of and $(a|b)\lor_\PS(c|d)$ is similar. They seem
complicated, but can be simplified, what we do below, and the Moore
machine representations are again much simpler and easier to analyze.

\begin{figure}[!hp]

\[\UseTips
\xymatrix  @C=20mm @R=12mm
{
&&*+++[o][F]{0}
\ar@(ul,ur)[]
\ar[d]^{\textstyle{dc^\C}}
\ar[rd]^{\textstyle{dc}}\\
{}\ar[r]&*+++[o][F]{0}
\ar@(u,l)[]
\ar[ur]^{\textstyle{ba}}
\ar[dr]_{\textstyle{dc}}
\ar@<1ex>[r]^(.65){\textstyle{ba^\C}}
\ar@<-1ex>[r]_(.65){\textstyle{dc^\C}}
\ar@/_3cm/[rr]_{\textstyle{abcd}}
&*+++[o][F]{0}
\ar@(ur,dr)[]
&*+++[o][F]{1}
\ar@(ur,dr)\\
&&*+++[o][F]{0}
\ar@(dl,dr)[]
\ar[u]_{\textstyle{ba^\C}}
\ar[ru]_{\textstyle{ba}}
}
\]

\caption[PS conjunction]{Moore machine of $\ps{(a|b)\land(c|d)}$}\label{f3}
\end{figure}

To simplify the $\TT$ representation above, let us define
$\first(a|b)$ to be  

\begin{equation}\label{first}\first(a|b):=(\PDIA(a\land b\land
\lnot\PREV\PDIA b))|\TRUE ).
\end{equation}

It is convenient to denote by $\first_\dwa(a|b)$ the first argument of
$\first(a|b).$

Then the conjunction can be easily and effectively described as
$(\first_\dwa(a|b)\land\first_\dwa(c|d)|\TRUE)$ (where $\land$ is the
classical conjunction of temporal logic).

The minimal Moore machine of $\first(a|b)$ is depicted in Fig.\
\ref{f4} below.

\begin{figure}[!hp]

\[\UseTips
\xymatrix @R=4mm
{
&&*+++[o][F]{1}
\ar@(ur,dr)[]\\
{}\ar[r]&*+++[o][F]{0}
\ar@(ur,dr)[]^{\textstyle{b^\C}}
\ar@/^/[ur]^{\textstyle{ba}}
\ar@/_/[dr]_{\textstyle{ba^\C}}\\
&&*+++[o][F]{0}
\ar@(ur,dr)[]
}
\]

\caption[PS simple conditional]{Moore machine of $\ps{(a|b)}=\first(a|b).$}\label{f4}
\end{figure}

\section{Embedding of existing {\em cea\/}'s and their incompleteness}

In this section we want to discuss the problem of embedding existing
\cea's into our model, and on that basis, the problem of defining
natural connectives among conditionals in general.

\subsection{Syntax}

We assume the following syntax of the {\em flat conditional
expressions}. The set of all such expressions will be denoted $\LL.$

The set of these expressions is the smallest set, containing all {\em
simple conditionals\/} of the form $(x|y),$ where $x,y\in\Omega,$ and
closed under two-ary (infix) operations $\land,\lor$ and one unary
prefix operation $\llnot.$

If we require the closure  under one additional binary operation
$(\cdot|\cdot)$ (which shouldn't be mixed up with the
parenthesis-bar-parenthesis construction appearing in simple
conditionals), we obtain the set of {\em full conditional expressions},
denoted $\LL^|.$

\subsection{Conditional event algebras}\label{cea}

According to \cite{kniga}, a {\em conditional event algebra\/} (\cea{}
in short) over a probability space $(\Omega,\fM,\Pr)$ is a space (but
not necessarily a probability space) $(\Omega_o,\fM_o,\Pr_o),$ extending
$(\Omega,\fM,\Pr),$ together with a function
$(\cdot|\cdot):\fM\times\fM\to\fM_o$ such that 
\begin{itemize}
\item $\fM_o$ is an algebra of the signature of boolean algebras.
\item $(a|b)=(a\cap b|b)$ for all $a,b\in \fM.$
\item The subalgebra of $(\Omega_o,\fM_o,\Pr_o)$ consisting of the
elements $(a|\Omega)$ is isomorphic to $(\Omega,\fM,\Pr)$ under the
bijection $a\mapsto(a|\Omega).$
\item $\Pr_o((a|b))=\Pr(a\cap b)/\Pr(b)$ for $a,b\in\fM,\ \Pr(b)>0.$
\item Certain equalities hold among the $\Pr_o$-probabilities.
\item The $\Pr_o$-probabilities for $\cap,\cup$ and ${}^\C$ of elements
of the form $(a_i|b_i)$ for $a_i,b_i\in\fM$ are effectively computable
from the set of $\Pr$-probabilities of all the boolean combinations of
the elements $a_i,b_i.$
\end{itemize}

\paragraph{How we understand \cea's.} It is readily seen, that any
particular \cea{} over any $(\Omega,\fM,\Pr)$ can be equivalently
considered as a mapping assigning elements of $\fM_o$ and probabilities
to flat conditional expressions. In this sense, \cea{} is a kind of a
{\em model\/} of a logic, whose syntax are the flat conditional
expressions. This is the way we understand \cea's and this is the level
on which we will criticize them.

\paragraph{Probabilistic models and \cea's.}

First of all, we do not think that the algebraic structure of a \cea{}
is particularly important. The language of flat conditional
expressions does not have equality, so what is really crucial are the
probabilities.  The algebraic structure can indeed be an obstacle
while assigning probabilities (this is perhaps why the authors of the
earlier papers devoted so much attention to it), but otherwise we are
not so much interested in it.

Next, what a \cea{} assigns to a conditional expression is definitely too
little. Apart from an element of $\fM_0$ and probability, an experiment
should be determined to verify experimentally the value of the
probability. This means, that the objects assigned to conditional
expressions should be events in a probabilistic space, i.e., the triple
$(\Omega_o,\fM_o,\Pr_o)$ should be indeed a probability space.
Moreover, the experiments for simple conditionals $(a|b)$ should
correspond to the natural experiments one performs to learn conditional
probabilities. In such experiments one can typically measure another
probabilistic parameters, like, e.g., the probability that $(a|b)$ is
defined. We think such additional parameters should be assigned to
compound conditional events, too.  Of course, in the existing algebras
we have hints concerning it, hidden in their universes and other details
of the constructions, provided by the inventors, but the very definition
of a \cea{} does not require the additional parameters to be even defined,
let alone to satisfy any reasonable properties. Strictly speaking, the
signature of \cea's is too small. In its present shape it permits
existence of other, isomorphic algebras, where all the additional
information is lost.

Last, but not least, in the applications of classical probability theory
one often encounters problems in modeling, typically of the following
form:  one has an event, whose meaning is completely clear (it is known,
when it happens and when it doesn't), but there is a problem of
specifying the probability space structure, and sometimes different
choices lead to different values for the probability of the same event.
In such circumstances one can only experimentally decide which of the
models is the correct one. A standard example is the difference between
the so called statistics of: Maxwell-Boltzmann, Bose-Einstein and
Fermi-Dirac, considered in quantum physics, and the unsuccessful search
for any elementary particle, which would satisfy the first statistics,
seemingly the most natural one among them (therefore we have bosons and
fermions, but we do not have maxwellons in physics) \cite{feller1}. The
tremendous success of probability theory in applications seems to
suggest this is the right way of creating a mathematical model of a real
life situation.  However, the \cea{} offers us another challenge. Even
after coming up with the proper model of probability space of
unconditional events, one still has a lot of work to choose the right
model for conditional events. 

In plain words, it means that the structure modeling conditional events
over a given probabilistic space of unconditional events, should be
functorial: given the former space, the space of conditional events
ought to be uniquely determined. Here the \cea's again fall short of
satisfying this requirement, because there are many known \cea's, and
each of them has its own definition of $(a|b)\land(c|d),$ with its own
probability, and all of them derive their definitions from certain first
principles. The answer is almost obvious --- the signature is too small,
and should consist of many different conjunctions, disjunctions and
negations, and perhaps lots of other connectives, which do not have any
natural counterparts in the nonconditional world.

\paragraph{$\TT$ as a solution.}

We believe that our system of temporal conditional events addresses all
of the problems we have indicated above.

Our conditional events belong to a normal probability space. They are
indeed stochastic processes --- projections of Markov chains, and all
their probabilistic properties (many more than just the bare
probability) can be verified experimentally, 

To the contrary, all the \cea's we call present tense in this paper have
a natural representation as algebras of $\trzy $-valued indicator
functions \cite[Chapter 3]{gnw91}.  Recalling that every event from
$\Sigma$ can be equivalently characterized by its $\dwa$-valued {\em
indicator random variable}, which is nothing but the characteristic
function, we should naturally expect that the lifting of $\Pr$ to the
space of $\trzy$-valued indicator functions, should consist of {\em
$\trzy $-valued random variables}, while the definition of a \cea{}
requires just elements of a strange algebra $\fM_o$ and numbers.  In
case of our temporal conditionals, we naturally expect conditionals to
be functions $\Omega^\infty\to \trzy ^\infty,$ and hence the lifting of
$\Pr$ to consist of random functions, which are nothing but {\em
stochastic processes}, the choice we actually have made.  This is the
natural pattern one should follow, and this is where the scalability of
the model is hidden. E.g., nothing could prevent us from considering
continuous time stochastic processes as models of conditional events, if
need be.

The space of conditional events is uniquely determined by the
probabilistic space of nonconditional events, and has many more natural
connectives than just three. In fact, we are not very original here:
already Schay \cite{MR37:5901} proposed a system with five connectives,
and considered adding even more of them.

Finally, to make our argument complete, we consider the major \cea's
below, and show that they embed in our models, in many different
ways. For the product space \cea, we employ the embeddings to
demonstrate that this \cea{} has some further defects. The existence
of multiple embeddings shows that the structure of a \cea's isn't
functorial just because there are many known \cea's, but because the
structure of a \cea{} does not prescribe the way the conditionals
relate to nonconditional events. Additionally, the product space
\cea{} is unable to characterize independence of conditional events by
means of equalities of probabilities.

\subsection{Present tense \cea's}

In our framework, there is a distinctive class of \cea's, which we call
{\em present tense \cea's.} The definitions of their connectives refer
to the present of the process, only, hence the name. Consequently, such
a connective applied to simple conditionals yields simple conditionals,
again.  Among such systems are SAC, proposed independently by Schay
\cite{MR37:5901}, Adams \cite{a86} and Calabrese \cite{c87} (who
extended it by an operator for re-conditioning), and GNW proposed by
Goodman \cite{g87} and Goodman, Nguyen and Walker \cite{gnw91} (later
Goodman and Nguyen \cite{gn96} proposed a re-conditioning operator for
GNW). They assume that all boolean combinations of simple conditional
expressions yield simple conditional expressions again. The definitions
of their connectives are given in Section \ref{teraz}. They do form
\cea's. Equivalently, \cite[Chapter 3]{gnw91}, these algebras can be
characterized as algebras of $\trzy$-valued indicator functions, and
their connectives are then characterized by mappings from certain
Cartesian power of $\trzy$ into $\trzy.$

We take the second point of view, and define their semantics as
follows. Our definition assigns to every conditional expression $e$ a
$\trzy$-valued indicator function $\sac{e}$ and $\gnw{e},$ respectively.
For us, indicator functions are nothing else than present tense
conditionals. (Originally these algebras do not involve time.) So, we
give the definition by describing translations
$\sac{\cdot},\gnw{\cdot}:\LL\to\CC.$ We use in the translations present
tense connectives of conditionals defined in \eqref{TLTL}.

The definition of SAC:

\begin{equation}\label{sac}
\begin{split}
\sac{(a|b)}&=(a|b),\\
\sac{e\land e'}&=\sac{e}\land_\SAC\sac{ e'},\\
\sac{e\lor e'}&=\sac{e}\lor_\SAC\sac{ e'},\\
\sac{\llnot e}&=\llnot_0\sac{e},\\
\sac{(e| e')}&=(\sac{e}|_\SAC\sac{ e'}).
\end{split}
\end{equation}

The definition of GNW:
\begin{equation}\label{gnw}
\begin{split}
\gnw{(a|b)}&=(a|b),\\
\gnw{e\land e'}&=\gnw{e}\land_\GNW\gnw{ e'},\\
\gnw{e\lor e'}&=\gnw{e}\lor_\GNW\gnw{ e'},\\
\gnw{\llnot e}&=\llnot_0\gnw{e},\\
\gnw{(e| e')}&=(\gnw{e}|_\GNW\gnw{ e'}).
\end{split}
\end{equation}

Given a probability space $(\Omega,\PP(\Omega),\Pr),$ the probability of
a conditional expression $e$ is
$\Pr_\SAC(e)={\Pr(\sac{e}=1)}/{\Pr(\sac{e}\neq \bot)},$ and
similarly $\Pr_\GNW(e)={\Pr(\gnw{e}=1)}/{\Pr(\gnw{e}\neq \bot)},$
provided that the denominators are nonzero.

All theses systems are readily seen to embed in our system of
conditionals.  In fact, if one represents them in the form of reduction
rules, as in \eqref{TLTL}, they do even embed syntactically in the $\TT$
logic. They are present tense because they do not contain temporal
connectives.

\subsection{Product space \cea}

However, there is another \cea, called the {\em product space} \cea,
which is not present tense. In order to analyze it and show that it can
be interpreted in our model, we have to give the definition.

The semantics is as follows:

Beginning with $(\Omega,\fM,\Pr),$ we form its countable power
$\Omega^\infty$ endowed with the product measure. The cylinder
$\underbrace{b\times\dots\times b}_j\times
a\times\Omega\times\Omega\times\cdots\subseteq \Omega^\infty$ for
$a,b\in\Sigma$ is denoted $b^j\times a\times\hat{\Omega}.$

Define the semantics function $\ps{\cdot}:\PS\to\PP(\Omega^\infty)$ by

\begin{align*}
\ps{(a|_\PS b)}&=
\bigcup_{i=0}^\infty (\Omega\setminus b)^j\times (b\cap
a)\times\hat{\Omega},\\
\ps{ e\land  e'}&=\ps{ e}\cap\ps{ e'},\\
\ps{ e\lor  e'}&=\ps{ e}\cup\ps{ e'},\\
\ps{\llnot  e}&=\Omega^\infty\setminus\ps{ e}.
\end{align*}

$\fM_o$ of the product space \cea\ is then the subalgebra of the
(boolean) algebra $\langle\PP(\Omega^\infty),\cup,\cap,
(\Omega^\infty\setminus \cdot)\rangle,$ generated by all elements
$b^j\times a\times\hat{\Omega}$ where $a,b\in\Sigma,$ and $\Pr_o$ is the
product measure.

There are indeed two versions of $\PS$: one defined in the paper
\cite{g94}, where equality of two conditionals is understood as true
equality of sets, and another, defined in \cite{kniga}, where the
equality of conditionals is understood as {\em equality almost
everywhere}, i.e., two conditional events of $\PS$ are equal iff their
symmetric difference has probability $0$. The latter is therefore not
logical, since it depends on the particular probability space structure.

The probabilities assigned to the elements of $\PS$ are those according
to the infinite product of $\Pr.$

\subsection{First embedding}\label{FE}

In order to construct the first embedding of $\PS$ into $\CC$ by
defining two operations $\ttl{\cdot}:\LL\to\TL$ and
$\ttt{\cdot}:\LL\to\TT$ as follows:

\begin{equation}\label{pierwszy}
\begin{split}
\ttl{(a| b)}&=\first_\dwa(a|b)\\
\ttl{e \land e'}&=\ttl{e}\land\ttl{e'}\\
\ttl{e \lor e'}&=\ttl{e}\lor \ttl{e'}\\
\ttl{\llnot e}&=\lnot\ttl{e}\\
\ttt{e}&=(\ttl{e}|\TRUE).
\end{split}
\end{equation}

$\ttt{e}$ (or, more formally, the conditional from $\CC$ represented by
the former) is the desired embedding.

\begin{lemma}\label{cztery-lematy}
For every expression $e\in\LL$

\begin{enumerate}
\item For every word $w\in\Omega^\infty$ holds
$(\ttt{e})_\infty(w)\in\dwa^\infty;$ 

\item Suppose $(a_1|b_1),\dots,(a_n|b_n)$ are all simple conditionals
occurring in $e.$ Suppose $w\in\Omega^\infty$ is so that
$b_{i_1},\dots,b_{i_k}$ are all events among $b_1,\dots,b_n$ which
happen in the sequence $w,$ and all of them happen not later than at
time $m.$ Then of the word $(\ttt{e})_\infty(w)$ is constant beginning
since time $m.$

\item $\ps{e}=\{w\in\Omega^\infty~/~\text{\rm $(\ttt{e})_\infty(w)$ is
eventually constant 1}\}$;

\item  $\Pr_o(e)=\Pr(\ttt{e})$;
\end{enumerate}
\end{lemma}
\begin{proof} The proof of 1., 2.\ and 3.\ goes by simultaneous
induction w.r.t.\ $e.$ For $e=(a| b)$ they follows from
a simple analysis of the definition of $\ps{e}$ and $\first_\dwa(a|b).$

Now consider $e=\lnot e'$ and assume by induction that 1., 2.\ and 3.\
hold for $e'.$ Since $\ttl{\lnot e}=\lnot\ttl{e'},$ we have 1.\ and 2.\
immediately.

Moreover, $(\lnot\ttl{e'}|\TRUE)_\infty(w)$ is eventually constant 1 iff
$(\ttl{e'}|\TRUE)_\infty(w)$ is eventually constant $0,$ which by 1.\
and 2.\ for $e'$ is equivalent to the fact that
$(\ttl{e'}|\TRUE)_\infty(w)$ is not eventually constant 1. This
concludes the induction step of 3.

Induction steps for the other connectives are equally simple.

We turn now to 4. Suppose $(a_1|b_1),\dots,(a_n|b_n)$ are all simple
conditionals occurring in a conditional expression $\varphi.$ W.l.o.g.\
assume $\Pr(b_i)>0$ for $i=1,\dots, k$ and $\Pr(b_i)=0$ for
$i=k+1,\dots, n.$ (We permit $k=0$ and $k=n,$ in which cases either all
$b_i$ are impossible, or all of them have positive probability.)

Represent $\Omega^\infty$ as a disjoint union of sets

\[A_n:=\{w\in\Omega^\infty~/~\begin{array}{c}
\text{$b_1,\dots,b_k$ happen in
$w$ and the first time}\\\text{when they all have already happened is
$n$}\end{array}\}\]

and the set $A_\infty:=\{w\in\Omega^\infty~/~\text{not all of
$b_1,\dots,b_k$ happen in $w$}\}.$ All these sets are clearly
measurable.

It is not hard to see that the (product) probability of
$A_\infty$ is $0,$ so
$\Pr_o(e)=\sum_{\substack{n\in\N\\A_n\subseteq\ps{e}}}
\Pr(A_n).$

It is not hard to verify, either, that

\[\lim_{n\to\infty}\sum_{m\in\N} \Pr\nolimits_n(A_m) \ =\ 1.\]

These two equalities imply 4.\ immediately, since for every $n$ 

\[\sum_{\substack{m\in\N\\A_m\subseteq\ps{e}}}
\Pr\nolimits_n(A_m)\leq \Pr\nolimits_n(\ttt{e})\leq 1-
\sum_{\substack{m\in\N\\A_m\cap\ps{e}=\varnothing}}
\Pr\nolimits_n(A_m).\]
\end{proof}

The following theorem follows now instantly.

\begin{theorem}\label{zaba} $\ttt{\cdot}$ is an embedding of the $\PS$
\cea{} into $\CC,$ in the sense that for any underlying probability
space, and any conditional expressions $e,e'$,  $\ttt{e}=\ttt{e'}$
iff $\ps{e}=\ps{e'},$ and $\Pr(\ttt{e})=\Pr_o(e).$\qed
\end{theorem}

\subsection{Reverse embeddings}

\begin{definition}
The reverse of a word $w=\omega_1\dots\omega_n\in \Omega^+,$ denoted
$w^{\RR},$ is $\omega_n\omega_{n-1}\dots \omega_1.$

Now consider a conditional $c\in\CC.$ Then $c^{\RR}\in\CC$ is a conditional 
defined by 

\[c^{\RR}(w):=c(w^{\RR}).\]
\end{definition}

The class of languages definable in $\TL$ is reverse-closed
\cite{Emerson}, i.e., if $L=\{w~/~w,|w|\models\varphi\}$ for some
$\varphi\in\TL$, then $L^{\RR}=\{w^{\RR}~/~w\in
L\}=\{w~/~w,|w|\models\psi\}$ for some$\varphi\in\TL$.  It follows
that $c^{\RR}$ is indeed a conditional in our sense.

\begin{theorem}\label{reverse}
For every $\star\in\trzy$
\[
\Pr(\ce{c}_n=\star)=\Pr(\ce{c^{\RR}}_n=\star).
\]

Consequently, $\Pr_o(c)=\Pr(c^\RR).$
\end{theorem}
\begin{proof}
$\Pr$ is understood here as a product measure, which is insensitive to
the order of its coordinates.
\end{proof}

It follows that any \cea, which can be at all isomorphically embedded
in our stochastic process model, has at least {\em two\/} embeddings,
which are reverses of each other. The only exception is when the
embedding is invariant under reverse, i.e., when each conditional $c$
in the image of the embedding satisfies $c(w)=c(w^\RR)$ for all
$w\in\Omega^+.$ However, it seems unlikely that any reasonable
embedding has this property. In particular, the natural embeddings of
the \cea's we consider here are not of this kind.

As a matter of example, we consider here $\PS$.  For the $\PS$
conjunction, its informal description of its reverse representation in
$\CC$ is that it is always defined and true iff the most recent defined
values of its both arguments were $1.$

In $\TT,$ we have that $\ttr{(a|b)\land(c|d)}$ (the reverse of the
embedding $\ttt{\cdot}$ defined in \eqref{pierwszy}) is defined by
$((b^\C\Since (a\land b))\land(d^\C\Since (c\land d))|\TRUE),$ whose
Moore machine is depicted below.

\begin{figure}[!hbp]

\[\UseTips
\xymatrix @C=16mm @R=36mm
{
&&*+++[o][F-]{0}
\ar@(ul,ur)
\ar@<2ex>[dll]|(.36){c^\C d\land (ab)^\C}
\ar@<2ex>[ddl]|(.36){abc^\C d}
\ar@<2ex>[dr]|(.36){ab\land (c^\C d)^\C}
\\
*+++[o][F-]{0}
\ar@(l,u)
\ar@<2ex>[rrr]|(.36){abcd}
\ar@<2ex>[rru]|(.36){cd\land (ab)^\C}
\ar@<2ex>[dr]|(.36){ab\land (cd)^\C}
&&&*+++[o][F-]{1}
\ar@(ur,dr)
\ar@<2ex>[lll]|(.36){a^\C bc^\C d}
\ar@<2ex>[lld]|(.36){c^\C d\land (a^\C b)^\C}
\ar@<2ex>[lu]|(.36){a^\C b\land (c^\C d)^\C}
\\
{}\save[]+<-2cm,3cm>\ar[u]\restore 
&*+++[o][F-]{0}
\ar@(rd,dl)
\ar@<2ex>[rru]|(.36){cd\land(a^\C b)^\C}
\ar@<2ex>[uur]|(.36){a^\C bcd}
\ar@<2ex>[ul]|(.36){a^\C b\land (ab)^\C}
}
\]

\caption[Reverse conjunction]{Moore machine of $\ttr{(a|b)\land(c|d)}.$}
\end{figure}

The precise definition of the reverse embedding of $\PS$ is as follows:
first, we take the original conditional expression
$e=e((a_1|b_1),\dots,(a_m|b_m))$ and replace every $(a_i|b_i)$ occurring
in it by $b_i^\C\Since (a_i\land b_i),$ obtaining $e'\in \TL,$ and then define
$\ttr{e}:=(e'|\TRUE).$ Formally:

\begin{equation}\label{drugi}
\begin{split}
\ttlr{(a| b)}&=b^\C\Since(a\land b)\\
\ttlr{e \land e'}&=\ttlr{e}\land\ttlr{e'}\\
\ttlr{e \lor e'}&=\ttlr{e}\lor \ttlr{e'}\\
\ttlr{\llnot e}&=\lnot\ttlr{e}\\
\ttr{e}&=(\ttlr{e}|\TRUE).
\end{split}
\end{equation}

For this particular embedding, we have the following consequence of
Theorem \ref{reverse} (and the simple fact that reversing is an
automorphism of the whole $\TT$)

\begin{theorem}\label{dwie-zaby} $\ttr{\cdot}$ is an embedding of
the $\PS$ \cea{} into $\CC,$ in the sense that for any underlying
probability space, and any conditional expressions $e,e'$,
$\ttr{e}=\ttr{e'}$ iff $\ps{e}=\ps{e'},$ and
$\Pr(\ttr{e})=\Pr_o(e).$\qed
\end{theorem}

\subsection{Sparse reverse embedding}

We give a new, radically different interpretation of $\PS$ in $\CC.$ The
main difference is that it is not an embedding. We are going to present
a way to interpret $\PS$ expressions in $\CC$ so that, for any
probability space $(\Omega,\PP(\Omega),\Pr),$ the $\Pr_o$-probability of
a $\PS$-expression is equal to the asymptotic probability of its
interpretation. However, expressions which yield equal element of the
$\PS$ \cea, may well give distinct conditional events in $\CC,$
although, as said before, these expressions will have equal asymptotic
probabilities.

First of all , we redefine the meaning of simple conditionals $(a|b).$
According to the new embedding, they are represented by $\TT$ formula

\[(a\land\lnot(\PBOX\lnot b\,\lor\,\lnot\PREV\TRUE)|b\lor \lnot\PDIA
b).\]

Denote this formula by $(a|_Sb).$ 

It is essentially the simple conditional $(a|b)\in\TT,$ except that it
is defined and false until $b$ becomes true for the very first time, and
since then behaves exactly like $(a|b)$ does in $\TT$.  The manipulation
is necessary to accommodate the $\PS$ principle, that degenerate simple
conditionals, like $(a|0),$ do have probability, and that it is $0.$

We now define the sparse reverse interpretation of $\PS$ in $\CC$ as
follows:  First, we take the original conditional expression
$e=e((a_1|b_1),\dots,(a_m|b_m))$ and set
$\ttrs{e}=(\ttlr{e}|\Lor_{i=1}^m(b_i\lor \lnot\PDIA b_i)),$ where
$\ttlr{\cdot}$ has been defined in \eqref{drugi}.

The conjunction and disjunction are defined precisely when at least
one of the arguments is defined, so they resemble the connectives of
$\SAC$ in this respect, but instead of assigning the other arguments
default values when they are undefined, like $\SAC$ does, their most
recent defined values are always used, instead.  Here is an example
Moore machine, in which we use $\epsilon$-moves. The graphical
representation in Fig. \ref{f00} and Fig. \ref{f0} shows that the new
operation is closely related to the reversed product space
conjunction, as can be expected from the shape of the $\TT$
representation.

\begin{figure}[htp]

\[
\xymatrix  @C=16mm @R=36mm
{
&&*+++[o][F]{0_{1\bot}}
\ar@(ur,ul)[]
\ar[r]_{a^\C bd^\C}
&*+++[o][F]{0_{0\bot}}
\ar@(ur,ul)[]
\ar@<-2ex>[l]_{abd^\C}
\\
{}\ar[r]&*+++[o][F]{0_{\bot\bot}}
\ar@(ul,u)[]
\ar[ur]|{abd^\C}
\ar[urr]|{a^\C bd^\C}
\ar[dr]|{b^\C cd}
\ar[drr]|{b^\C b^\C d}
\\
&&*+++[o][F]{0_{\bot1}}
\ar@(dr,dl)[]
\ar[r]^{b^\C c^\C d}
&*+++[o][F]{0_{\bot0}}
\ar@(dr,dl)[]
\ar@<2ex>[l]^{b^\C cd}
}
\]

\caption[Sparse reverse conjunction 1.]{Moore machine of
$\ttrs{(a|b)\land(c|d)},$ part 1. This part of the machine is the
transient part of the Markov chain, when $\Pr(b),\Pr(d)>0$ (Lemma
\ref{12}) and the whole reachable part when at least one of these
probabilities is $0$ (Lemma \ref{11}).\\ Subscripts of the state
labels indicate the most recent value of $(a|b)$ and $(c|d),$
respectively.}\label{f00}
\end{figure}

\begin{figure}[htp]

\[\UseTips
\xymatrix @C=16mm @R=36mm
{
{}\save[]+<0cm,-18mm>*+++[o][F-]{\bot}
\ar@/^/[d]^{\epsilon}
\ar@/_/@{<-}[d]_{b^\C d^\C}
\restore
&&*+++[o][F-]{0_{01}}
\ar@(ul,ur)
\ar@<2ex>[dll]|(.36){c^\C d\land (ab)^\C}
\ar@<2ex>[ddl]|(.36){abc^\C d}
\ar@<2ex>[dr]|(.36){ab\land (c^\C d)^\C}
\ar@/^/[r]^{b^\C d^\C}
&*+++[o][F-]{\bot}
\ar@/^/[l]^{\epsilon}
\\
*+++[o][F-]{0_{00}}
\ar@(l,ul)
\ar@<2ex>[rrr]|(.36){abcd}
\ar@<2ex>[rru]|(.36){cd\land (ab)^\C}
\ar@<2ex>[dr]|(.36){ab\land (cd)^\C}
&&&*+++[o][F-]{1_{11}}
\ar@(ur,r)
\ar@<2ex>[lll]|(.36){a^\C bc^\C d}
\ar@<2ex>[lld]|(.36){c^\C d\land (a^\C b)^\C}
\ar@<2ex>[lu]|(.36){a^\C b\land (c^\C d)^\C}
\\
*+++[o][F-]{\bot}
\ar@/_/[r]_{\epsilon}
&*+++[o][F-]{0_{10}}
\ar@(rd,dl)
\ar@<2ex>[rru]|(.36){cd\land(a^\C b)^\C}
\ar@<2ex>[uur]|(.36){a^\C bcd}
\ar@<2ex>[ul]|(.36){a^\C b\land (ab)^\C}
\ar@/_/[l]_{b^\C d^\C} &
&{}\save[]+<0cm,18mm>*+++[o][F-]{\bot}
\ar@/^/[u]^{\epsilon}
\ar@/_/@{<-}[u]_{b^\C d^\C}
\restore
}
\]

\caption[Sparse reverse conjunction 2.]{Moore machine of $\ttrs{(a|b)\land(c|d)},$ part 2. This part
of the machine is the (only) ergodic class of the Markov chain, when
$\Pr(b),\Pr(d)>0$ (Lemma \ref{12}), and is unreachable  when at
least one of these probabilities is $0$ (Lemma \ref{11}).\\ Subscripts
of the state labels indicate the most recent value of $(a|b)$ and
$(c|d),$ respectively.} \label{f0}
\end{figure}

\begin{figure}[htp]
\[
\begin{array}{|c|c|c|}
\hline
\multicolumn{3}{|c|}{\text{Missing arrows}}\\
\hline\hline
\text{From}&\text{To}&\text{Label}\\
\hline
0_{\bot\bot},0_{\bot0},0_{\bot1},0_{1\bot},0_{0\bot}&0_{00}&a^\C bc^\C d
\\
\hline
0_{\bot\bot},0_{\bot0},0_{\bot1},0_{1\bot},0_{0\bot}&0_{01}&a^\C bcd
\\
\hline
0_{\bot\bot},0_{\bot0},0_{\bot1},0_{1\bot},0_{0\bot}&0_{10}&abc^\C d
\\
\hline
0_{\bot\bot},0_{\bot0},0_{\bot1},0_{1\bot},0_{0\bot}&1_{11}&abcd
\\
\hline
0_{1\bot}&1_{11}&b^\C cd
\\
\hline
0_{0\bot}&0_{01}&b^\C cd
\\
\hline
0_{1\bot}&0_{10}&b^\C c^\C d
\\
\hline
0_{0\bot}&0_{00}&b^\C c^\C d
\\
\hline
0_{\bot1}&1_{11}&abd^\C
\\
\hline
0_{\bot0}&0_{10}&abd^\C
\\
\hline
0_{\bot1}&0_{01}&a^\C bd^\C
\\
\hline
0_{\bot0}&0_{00}&a^\C bd^\C\\
\hline
\end{array}
\]
\caption[Sparse reverse conjunction 3.]{Moore machine of $\ttrs{(a|b)\land(c|d)},$ part 3. This part
of the machine is the table of the transitions from the ``transient''
part (Fig.~\ref{f00}) to the ``ergodic'' part
(Fig.~\ref{f0}).}\label{f000}
\end{figure}

\begin{lemma}\label{11}
If $\Pr(b_i)=0$ for at least one $1\leq i\leq m,$ then
$\Pr(\ttrs{e})=\Pr(\ttt{e}).$
\end{lemma}
\begin{proof} In this case, assuming $\Pr(b_i)=0$, we have that
$\Lor_{i=1}^m(b_i\lor \lnot\PDIA b_i))$ is true with probability $1,$
hence $\ce{\ttrs{e}}=\ce{\ttr{e}}$ with probability $1.$
Now the thesis follows immediately from Theorem \ref{dwie-zaby}.
\end{proof}

\begin{lemma}\label{12} If $\Pr(b_i)>0$ for all $1\leq i\leq m,$ then
$\Pr(\ttrs{e})=\Pr_o(\ttt{e}).$
\end{lemma}
\begin{proof} Let $e=e((a_1|b_1,\dots,(a_n|b_n)).$ We prove
$\Pr(\ttr{e})=\Pr(\ttrs{e}),$ which is, by Theorem
\ref{dwie-zaby}, equivalent to what we have to show.

Denote $t=$ the first moment $m$ when all the $b_i$'s have already
been defined.

Let us note that, if $t<n,$ then the event
$\ce{\ttrs{e}}_n=\bot$ is independent of the whole history of
$\ce{\ttrs{e}}$ up to time $n-1.$ This is so because the decision
whether $\ce{\ttrs{e}}_n$ is defined or not depends solely on the
present time values of $b_i$'s, and their present time values become
independent of the (strict) past, when the condition
$\Lor_{i=1}^n\lnot\PDIA b_i$ becomes for the first time false,
because it remains then false forever, and the ``given'' part of
$\ttrs{e}$ does not contain any other time modalities.

Moreover, whenever
$\ce{\ttrs{e}}(\omega_1\dots\omega_n)\neq\bot,$ then in fact
$\ttrs{e}(\omega_1\dots\omega_n)=
\ttr{e}(\omega_1\dots\omega_n),$ which is clear from the
syntactic representation of both conditional objects.

Denote for convenience

$q=\Pr(\ce{\ttrs{e}}_n=\bot|t<n)=1-\Pr(\Lor_{i=1}^m b_i)<1,$
as well as and $c=\ttr{e}$ and $c^\S=\ttrs{e}.$

Fix $\varepsilon>0.$ Let $M$ be large enough to have $\Pr(t\geq
M)<\varepsilon.$ Let $N$ be a large integer, and let $n$ satisfy
$n-N>M.$

We have then by the independence

\begin{align*}
\Pr(\ce{c}_n=1)&\geq
\sum_{i=0}^N\Pr(\ce{c^\S}_{n-i}=1)\Pr(\ce{c^\S}_{n-i+1}=\bot)
\dots \Pr(\ce{c^\S}_{n}=\bot)-\varepsilon,\\
\Pr(\ce{c}_n=0)&\geq
\sum_{i=0}^N\Pr(\ce{c^\S}_{n-i}=0)\Pr(\ce{c^\S}_{n-i+1}=\bot)
\dots \Pr(\ce{c^\S}_{n}=\bot)-\varepsilon,
\end{align*}

where the $\varepsilon$ error terms are caused by the event that $t\geq M.$ 

Because each of the $\Pr(\dots)$ expressions above tends to a limit as
$n$ approaches infinity, we get 

\begin{align*}
\Pr(c)&\geq
\sum_{i=0}^N\lim_{n\to\infty}\Pr(\ce{c^\S}_{n-i}=1)\lim_{n\to\infty}\Pr(\ce{c^\S}_{n-i+1}=\bot)
\cdots \Pr(\ce{c^\S}_{n}=\bot)-\varepsilon,\\
&=\sum_{i=0}^N\lim_{n\to\infty}\Pr(\ce{c^\S}_{n}=1)\lim_{n\to\infty}\Pr(\ce{c^\S}_{n}=\bot)
\cdots \Pr(\ce{c^\S}_{n}=\bot)-\varepsilon\\
&=\sum_{i=0}^N\lim_{n\to\infty}\Pr(\ce{c^\S}_{n}=1)(\lim_{n\to\infty}\Pr(\ce{c^\S}_{n}=\bot))^i-\varepsilon\\
&=\sum_{i=0}^N\lim_{n\to\infty}\Pr(\ce{c^\S}_{n}=1)q^i-\varepsilon\\
&=\frac{1-q^N}{1-q}\lim_{n\to\infty}\Pr(\ce{c^\S}_{n}=1)-\varepsilon,
\end{align*}

and similarly 

\[1-\Pr(c)\geq
\frac{1-q^N}{1-q}\lim_{n\to\infty}\Pr(\ce{c^\S}_{n}=0)-\varepsilon.
\]

In the limit $N\to\infty$ both inequalities become

\begin{align*}
\Pr(c)&\geq
\frac{1}{1-q}\lim_{n\to\infty}\Pr(\ce{c^\S}_{n}=1)-\varepsilon\\
1-\Pr(c)&\geq
\frac{1}{1-q}\lim_{n\to\infty}\Pr(\ce{c^\S}_{n}=0)-\varepsilon,
\end{align*}

hence 

\begin{align*}
\Pr(c)&\geq
\frac{1}{1-q}\lim_{n\to\infty}\Pr(\ce{c^\S}_{n}=1)-\varepsilon\\ 
\Pr(c)&\leq
1-\frac{1}{1-q}\lim_{n\to\infty}\Pr(\ce{c^\S}_{n}=0)+\varepsilon\\
&=\frac{1}{1-q}(1-q -\lim_{n\to\infty}\Pr(\ce{c^\S}_{n}=0))+\varepsilon\\
&=\frac{1}{1-q}\lim_{n\to\infty}\Pr(\ce{c^\S}_{n}=1)+\varepsilon,
\end{align*}

because
$1=\Pr(\ce{c^\S}_{n}=1)+\Pr(\ce{c^\S}_{n}=0)+\Pr(\ce{c^\S}_{n}=\bot),$
in which  the last term is constant equal $q.$

We took an arbitrary $\varepsilon>0,$ therefore  indeed

\[
\Pr(c^\S)=\frac{1}{1-q}\lim_{n\to\infty}\Pr(\ce{c^\S}_{n}=1),
\]

and likewise 

\[
1-\Pr(c^\S)=\frac{1}{1-q}\lim_{n\to\infty}\Pr(\ce{c^\S}_{n}=0).
\]

{}From the last two equalities the equality $\Pr(c)=\Pr(c^\S)$
follows immediately.
\end{proof}

Summing up,

\begin{theorem}\label{sparse} For every conditional expression $e$ and
every probability assignment to the elements in $\Omega,$
$\Pr_o(e)=\Pr(\ttrs{e}).$\qed
\end{theorem}

The very important consequence of the theorem is the following:

\begin{corollary} The formalism of $\PS$ \cea{}, seen as a logic of
conditionals, is unable to determine certain probabilistic
characteristics, other than the asymptotic probability, associated with
stochastic processes.
\end{corollary}
\begin{proof} $\PS$ possesses two interpretations in $\CC,$ one of which
consists entirely of always defined temporal conditionals (the
$\ttt{\cdot}$ embedding), while the second contains conditionals which
are defined with asymptotic probability strictly less than 1 (the
$\ttrs{\cdot}$ interpretation).\end{proof}

Another similar example of deficiencies of the $\PS$ \cea{} can be found
below, Theorem \ref{(a|0)}.

Let us note that the $\ttrs{\cdot}$ interpretation does not preserve
the algebraic structure of the $\PS$ \cea, in general. Indeed, already
$\ps{(0|a)}=\ps{(0|b)}$, while $\ttrs{(0|a)}=(a|_Sb)$ and
$\ttrs{(0|b)}=(0|_Sb)$ represent different conditional objects, for
$a\neq b,$ $a,b\in\E.$

\section{Advantages of $\TT$ conditionals}

\subsection{Complexity and proof systems}

The paper \cite{g94} asks for the proof systems for various
three-valued logics, appearing in the context of the theory of
conditionals. 

In order to discuss this issue, we use the machinery of complexity
theory. All the necessary definitions can be found in \cite{HU}.

As we prove below, for all of the major \cea's, the sets of weak
tautologies are $\coNP$ complete.

In the light of the above results, there is a little hope for a
practically useful proof system for the most prominent systems among
$\GNW$ and $\SAC.$ Indeed, unless $\NP=\coNP,$ a very unlikely
complexity-theoretic collapse, for {\em every\/} sound and complete
proof system for the two above logics, there must be weak tautologies of
length $n$ such that their shortest proofs are of superpolynomial length
w.r.t.\ $n,$ for infinitely many $n.$ 

Temporal logic is known to be $\PSPACE$-complete, Theorem \ref{Emerson}.
It is therefore obvious that the set of weak tautologies of $\TT,$
consisting of all expressions $(\varphi|\psi)$ such that for every
$w\in\trzy^+,$ $(\varphi|\psi)(w)\in\{1,\bot\}$ (equivalently: that
$\psi\to\varphi$ is a tautology of $\TL$), is $\PSPACE$-complete, too.

Although it is commonly believed that $\coNP\subsetneq\PSPACE,$ from
practical standpoint both admit exponential time algorithms, and no
better ones are known. Consequently, the practical algorithmic
difference between \cea's and $\TT$ is not so crucial. The advantage of
considering \cea's as subsystems of $\TT$ stems from the fact that a lot
is known about proof systems for temporal logic --- unlike for \cea's.

We are not interested in the complexity of \cea's as logics involving
terms $(a|b)$ as atoms, but rather as $\trzy$-valued logics. To explain
the difference, let us note that in the calculus of any of the \cea's
one can easily restrict atoms to be two-valued (e.g., by using only
atoms of the form $(a|1)$), and thus all the complexity questions
trivialize, the sets of tautologies of all the logics are
$\coNP$-complete.  Here, we assume that the atoms can always assume all
three logical values, and are effectively variables.  Hence we indeed
view our (weak) tautologies as a kind of (weak) meta-tautologies, i.e.,
formulas which evaluate to either $1$ or $\bot,$ no matter what the
arguments are.

Formally, for this section we modify $\LL$ and $\LL^|$ replacing simple
conditionals $(a|b)$ by variables $p_1,p_2,\dots$ in conditional
expressions.

Likewise, given a valuation $v:\{p_1,p_2,\dots\}\to\trzy,$ we let
$\sac{\cdot}^v$ and $\gnw{\cdot}^v$ assign values in $\trzy$ to
expressions in $\LL^|$. These values are determined by the equations in
\eqref{sac} and \eqref{gnw}.

An expression $e$ is a {\em weak tautology\/} of $\SAC$ \cea{} ($\GNW$
\cea, respectively) iff $\sac{e}^v\neq 0$ ($\gnw{e}^v\neq 0,$
respectively) for every $v$.

$e$ is a strong tautology of $\SAC$ ($\GNW$, respectively), iff the
above values are 1 for every $v.$

We consider the complexity problem of determining if an expression $e$
is a weak tautology according to each of the considered \cea's,
considering also some syntactical restrictions put on the syntactical
shape of $e.$

We do not consider strong tautologies, which is explained by the
following.

\begin{proposition} In $\GNW$\/ and $\SAC$\/ there are no strong
tautologies.
\end{proposition}
\begin{proof} All connectives  have value $\bot$ if all their arguments
are $\bot,$ for both $\SAC$ and $\GNW.$
\end{proof}

Occasionally, we want to consider $\LL^|$ as the syntax of classical
logic. In this case, given a valuation $v:\{p_1,p_2,\dots\}\to\dwa,$ 
the value $\cl{e}^v\in\dwa$ is computed according to the classical
rules, where the conditioning $|$ is understood as reverse implication:
$(a|b)$ is $a\leftarrow b.$

\paragraph{Pure conditional parts.}

We consider here pure conditional fragments of $\SAC$ and $\GNW,$ i.e.,
expressions in which the only connective used is $|.$

It shows that unlimited use of re-conditioning leads to
$\coNP$-completeness of the weak tautology problem.

\begin{theorem}\label{pure|} It is an $\coNP$-complete problem to
determine if an expression $e\in\LL^|$ involving only re-conditioning is
a weak tautology of $\SAC.$

It is an $\coNP$-complete problem to determine if an expression
$e\in\LL^|$ involving only re-conditioning is a weak tautology of
$\GNW.$
\end{theorem}
\begin{proof}
It is obvious that the sets of weak tautologies in both cases are in
$\coNP.$ So it remains to prove their hardness in this complexity class.

It is easily seen that the (re-)conditioning operators of both $\GNW$
and $\SAC$ satisfy the following property: the equivalence relation
$\approx$ on $\trzy $ identifying $1$ with $\bot$ is a congruence of
$\A=\langle\trzy ,{|_\SAC}\rangle$ and $\B=\langle\trzy
,{|_\GNW}\rangle,$ and the quotient algebras both $\A=\langle\trzy
,{|_\SAC}\rangle/{\approx}$ and $\B=\langle\trzy
,{|_\GNW}\rangle/{\approx}$ are isomorphic to the 2-element algebra with
the reversed classical implication $\langle\dwa ,\leftarrow\rangle.$ The
natural epimorphism $\eta_\A:\A\to\langle\dwa ,\leftarrow\rangle$ sends 1 and
$\bot$ to 1, and 0 to 0, and the definition of $\eta_\B$ is identical.

Therefore a pure conditional expression is a weak tautology of either of
the considered \cea's iff it is a classical tautology, after its
(re-)conditioning operator is replaced by the reversed classical
implication. The classical formula resulting from this replacement is
denoted $\bar{e}.$

We have to prove that $e$ is not a weak  tautology iff $\bar{e}$
is not a tautology. Let $v$ be any valuation of the variables of
$e$ in $\trzy .$ Now we use the natural epimorphism $\eta_\A$
and get

\[\cl{\bar{e}}^{\eta_\A\circ v}=\eta_\A(\sac{e}^v).\]

So if one of the values above can be $0,$ the other can be, as well,
which establishes the desired equivalence.

Since it is known that the tautologies of the classical propositional
logic of pure implication are $\coNP$ complete \cite{H}, the claim
follows. 
\end{proof}

As a by-product we have 

\begin{corollary} The sets of pure conditional weak tautologies of
$\SAC$\/ and $\GNW$\/ are identical.\qed
\end{corollary}

\paragraph{Flat parts.}

Here we consider $\SAC$ and $\GNW$ without re-conditioning. 

We can define the following $\NP$-complete problem 3CNF-SAT.

Given: an expression $e\in \LL$ of the following syntactical form:

\begin{equation}\label{3CNF-SAT}
e=(\ell_{11}\lor \ell_{12}\lor \ell_{13})\land (\ell_{21}\lor
\ell_{22}\lor \ell_{23})\land \dots\land (\ell_{s1}\lor \ell_{s2}\lor
\ell_{s3}),\end{equation}

where each of the $\ell_{ij}$ is either $p_j$ or $\llnot p_j.$

The $\NP$-complete problem is: given $e$ of the above shape, determine
if $e$ is satisfiable, i.e, if there exists $v$ such that $\cl{e}^v=1.$

It follows that it is $\coNP$-complete to determine, given $e$ as above,
if $e$ is {\em not\/} satisfiable, i.e., whether $\cl{\llnot e}^v=1$ for
every $v.$

In order to prove $\coNP$-completeness of the sets of weak tautologies
of either of the \cea's, we have to construct a polynomial time
computable transformation $e\mapsto \bar{e}$ translating $e$ of the form
\eqref{3CNF-SAT} into $\bar{e}$ of the form conforming to the restriction set in
the respective theorem, and such that $\llnot e$ is not satisfiable in the
classical sense iff $\bar{e}$ is a weak tautology of the respective
logic.

\begin{theorem} It is an $\coNP$-complete problem to determine if an
expression $e\in\LL$ is a weak tautology of $\SAC.$

It is an $\coNP$-complete problem to determine if an expression
$e\in\LL$ is a weak tautology of $\GNW.$ 
\end{theorem}

\begin{proof} It is easily seen that the connectives $\land_\GNW$ and
$\lor_\GNW$ satisfy again the property that the equivalence relation
$\approx$ on $\trzy$ identifying $1$ with $\bot$ is a congruence of the
algebra with the above functions, and the quotient algebra
$\langle\trzy, \land_\GNW,\lor_\GNW\rangle\big/\approx$ is isomorphic to
the classical $\langle\dwa,\land,\lor\rangle.$ This fails about the
negation, however.

As the negation is applied to atoms only in 3CNF-SAT, we do not have to use
the negation of $\GNW$ directly.  Instead, we introduce new variables to
denote the negations, and force them to behave correctly outside of the
translation of $e.$

Formally, let the mapping $e\mapsto e'$ from the classical propositional
logic into $\GNW$ be defined by replacing unnegated atoms $p$ in $e$ by
$\hat{p}$ and negated atoms $\llnot p$ by $\check{p}.$ Concerning binary
connectives, we leave $\land$ and $\lor$ untouched.

Then let $\bar{e}$ be defined as $\llnot(e'\land{\Land\limits_p}
(\hat{p}\lor \check{p})\land(\llnot\hat{p}\lor\llnot \check{p})),$ where
$p$ in the big conjunction ranges over all propositional variables of
$e.$

Certainly the mapping $e\mapsto\bar{e}$ is computable in polynomial
time. In order to show the $\coNP$ completeness of the set of
tautologies of $\GNW,$ it suffices to show two implications:

\begin{itemize}
\item if $e$ is satisfiable classically, then $\bar{e}$ is not a weak
tautology of $\GNW.$ 
\item if then $\bar{e}$ is not a weak tautology of $\GNW,$ then $e$ is
satisfiable classically.
\end{itemize}

For the first item, assume that $e$ is satisfiable, i.e., there is an
assignment $v$ of $0$'s and $1$'s to the propositional variables of $e$
which makes $e$ into $1.$ We construct a $\trzy$-valued assignment $w$
which makes $\bar{e}$ into $0$. If $v(p)=1,$ we let $w(\hat{p})=1$ and
$w(\check{p})=0.$ If $v(p)=0,$ we let $w(\hat{p})=0$ and
$w(\check{p})=1.$ In $e'$ each variable has under $w$ exactly the value
of the corresponding literal in $e$ has under $v.$ So $e'$ evaluates to
$1,$ because connectives in $\GNW$ behave classically for classical
arguments. In addition, each of the formulas $(\hat{p}\lor
\check{p})\land(\llnot\hat{p}\lor\llnot\check{p})$ evaluates to $1,$ so
altogether $\bar{e}$ evaluates to the $\llnot_0$-negation of the value
to which $e'$ does evaluate, which is $0,$ as desired.

For the second item, assume there is an assignment $w$ of $0$'s, $1$'s
and $\bot$'s to the propositional variables of $\bar{e}$ which makes it
$0.$ It follows that each of the terms
$(\hat{p}\lor\check{p})\land(\llnot\hat{p}\lor\llnot\check{p})$ must
evaluate to $1$ under $w.$ Therefore of each pair $\hat{p},\check{p},$
one variable must be assigned $1$ and the other $0$ by $w,$ which can be
checked by simple inspection of all possibilities.  Moreover, $e'$ must
evaluate to $1$ under $w,$ which is indeed $\dwa$-valued, by the
previous observation. The connectives of $\GNW$ act classically for
classical arguments, therefore $e$ is indeed classically satisfiable, by
the valuation $v:p\mapsto w(\hat{p}).$ 

This finishes the proof.
\end{proof}

\begin{theorem} The weak flat-conditional $\SAC$ is $\coNP$-complete.
\end{theorem}
\begin{proof} We are going to use the same proof idea as before.
However, we have a small problem. The conjunction of $\SAC$ does not
permit us to deduce, that if a conjunction of two formulas evaluates to
$1,$ so does each of the components.  

So instead of the original conjunction, we have to use some custom
connective defined from the conjunction, disjunction and negation, which
will act as a ``good'' conjunction, for which the inference does hold.
It turn out, that the conjunction of $\GNW$ is not definable in $\SAC,$
but there is another connective we can use instead, and which is
definable (we discuss the definability of connectives in $\SAC$ and
$\GNW$ in another paper \cite{SAC-GNW}).  Its definition is as follows: 
\[x\sqcap y\equiv [ x\lor( y\land( x\lor\llnot y))]\land[ y\lor( x\land
(y\lor\llnot x))].\]

It is not difficult (but tedious) to check, that $\sac{x\sqcap y}$ has
truth table

\[\conn\sqcap00001000,\]

which is exactly what we need for our purposes. The only subtle point
is that our $x\sqcup y$ is substantially longer than $|x|+|y|.$ Indeed
it is about 4 times longer. We do replace $\land$ by $\sqcap$ in very
long conjunctions. However, if we represent this long conjunction as a
balanced binary (parse) tree, i.e., insert brackets to obtain the
structure
\[(((\ldots\sqcap\ldots)\sqcap(\ldots\sqcap\ldots))\sqcap((\ldots\sqcap\ldots)\sqcap(\ldots\sqcap\ldots))),\]
the depth of nesting of conjunctions is at most log base 2 of the number
$N$ of clauses in the conjunction, and the total increase of length
caused by the replacement is $4^{\text{depth of
nesting}}=4^{\log_2N}=N^2.$ Altogether, the resulting formula, using
$\sqcap$ in place of $\land,$ is still of polynomial size, and can be
easily constructed in polynomial time, as needed.
\end{proof}

\subsection{Independence of conditional events}

There has been a considerable amount of interest in the independence
issue for conditional events, reflected in the \cea{} literature
\cite{g94,c97,p88}. The problem is that typically even for $a$ and $b$
mutually independent of $c$ and $d$ one does not have
$\Pr((a|b)\land(c|d))=\Pr((a|b))\Pr((c|d)).$ The only exception is
$\PS,$ where this equality holds. The other variant of independence:
$\Pr((a|b)|(c|d))=\Pr((a|b))$ is undefined in some formalisms, due to
the lack of re-conditioning operator, and fails in others.  However,
note that in the \cea{} framework one cannot obtain any proper
characterization of independence, because there is no underlying
probabilistic semantics, in which one could say which pairs of
conditionals are independent and which aren't, and then attempt to
characterize this by equalities among probabilities.  One {\em feels\/}
that $(a|b)$ and $(c|d)$ should be independent for mutually independent
arguments, but this is not more than a feeling, and there is no idea
there what might make two conditionals independent when their arguments
are not mutually independent, or when they are composite. 

We can address this problem in our semantical setting.  First of all,
for $a$ and $b$ mutually independent of $c$ and $d,$ the stochastic
processes $\ce{(a|b)}$ and $\ce{(c|d)}$ are obviously independent.  And
of course, the {\em independence of the stochastic processes\/} is what
the independence of conditionals should be. This remains true, no matter
which \cea\ we consider. It is, however, a different story if this
independence can be formally characterized in terms of equalities
between probabilities of conditionals in the \cea\ under consideration.
It appears that in the pure \cea{} formalism this cannot be achieved,
because in Theorem \ref{PS-indep} below we show that independence is
undefinable in the $\PS$ \cea.

To be precise, the full independence of stochastic processes $\X,\Y$
means that {\em the full histories of both processes\/} are
independent, which is different from the much less restrictive requirement that
just the present time values should be independent. The first version
is formalized by the requirement that $\X_{+,t}$ and $\Y_{+,t}$ are
independent at any time $t>0,$ i.e., for any $w_1\dots w_t,v_1\dots
v_t\in\trzy^t$ holds 

\begin{multline*}
\Pr\left(\hspace{-6pt}
\begin{array}{ccc}
X_1=w_1,&\dots,&X_t=w_t\\
Y_1=v_1,&\dots,&Y_t=v_t
\end{array}
\hspace{-6pt}\right)=\\
\Pr(X_1=w_1,\dots,X_t=w_t)\Pr(Y_1=v_1,\dots,Y_t=v_t).
\end{multline*}

The weaker, present tense independence requires only that $X_t$ and $Y_{t}$
are independent at any time $t>0,$ i.e., that for any $w,v\in\trzy$ holds
$\Pr(X_t=w,Y_t=v)=\Pr(X_t=w)\Pr(Y_t=v).$ To see the difference it is worth
noting that for {\em any present tense\/} $\TT$ formula $(\varphi|\psi)$ the
processes $\ce{(\varphi|\psi )}$ and $\ce{(\PREV \varphi|\PREV \psi )}$ are
present tense independent, although of course they are easily seen to be
dependent, unless the former is constant.

But let us note the following simple fact.

\begin{lemma}\label{present-tense} If $c_1$ and $c_2$ are two present
tense conditionals, they are independent iff they are present tense
independent.\qed
\end{lemma}

We know now what independence should {\em mean.} It is another story how to
{\em characterize\/} it in terms of the asymptotic probability of
conditionals.

First we prove the characterization for present tense independence at fixed
time.

Let $\defn (a|b):=(b|\TRUE).$

\begin{lemma}\label{indep}
Let $n$ be a fixed time instant. The following  are equivalent:
\begin{itemize}
\item Random variables  $\ce{(a|b)}_n$ and $\ce{(c|d)}_n$ are 
independent.
\item The following four equalities hold:
\begin{align}
\pr_n((a|b)\land_\Sch(c|d))&=\pr_n((a|b))\pr_n((c|d))\label{i1}\\
\pr_n((a|b)\land_\Sch\defn(c|d))&=\pr_n((a|b))\pr_n(\defn(c|d))\label{i2}\\
\pr_n(\defn(a|b)\land_\Sch(c|d))&=\pr_n(\defn(a|b))\pr_n((c|d))\label{i3}\\
\pr_n(\defn(a|b)\land\defn(c|d))&=\pr_n(\defn(a|b))\pr_n(\defn(c|d))\label{i4},
\end{align} 
where we assume an equation to hold in case when both sides are
undefined.
\end{itemize}
\end{lemma}
\begin{proof} 
$\Downarrow$ Independence of random variables $\ce{(a|b)}_n$ and
$\ce{(c|d)}_n$ implies, in particular, that

\begin{equation}\tag{$\ref{i4}'$}\label{i4'}
\Pr(\ce{(a|b)}_n=0,1,\ce{(c|d)}_n=0,1)=
\Pr(\ce{(a|b)}_n=0,1)\Pr(\ce{(c|d)}_n=0,1),
\end{equation}

which is exactly equivalent to \eqref{i4}. The other consequences of
independence are equalities

\begin{align}
\Pr(\ce{(a|b)}_n=1,\ce{(c|d)}=1)&=\Pr(\ce{(a|b)}_n=1)\Pr(\ce{(c|d)}=1)
\tag{$\ref{i1}'$}\label{i1'}\\
\Pr(\ce{(a|b)}_n=1,\ce{(c|d)}_n=0,1)&=
\Pr(\ce{(a|b)}_n=1)\Pr(\ce{(c|d)}_n=0,1)
\tag{$\ref{i2}'$}\label{i2'}\\
\Pr(\ce{(a|b)}_n=0,1,\ce{(c|d)}_n=1)&=
\Pr(\ce{(a|b)}_n=0,1)\Pr(\ce{(c|d)}_n=1),
\tag{$\ref{i3}'$}\label{i3'}
\end{align}

which, divided by \eqref{i4'}, yield \eqref{i1}, \eqref{i2} and
\eqref{i3}, respectively. Note that if both sides of \eqref{i4'} are $0,$ then
all the resulting equalities involve an undefined term on both sides, and
hence hold, according to our convention.
\bigskip

$\Uparrow$ Let \eqref{i4'} (i.e., \eqref{i4}) hold. If its both sides are $0,$
the random variables $\ce{(a|b)}_n$ and $\ce{(c|d)}_n$ are independent,
because one of them is constant. So let us assume \eqref{i4'} holds and its
both sides are nonzero. In particular, each of the \eqref{i1}, \eqref{i2} and
\eqref{i3} is defined on both sides, because the denominators are everywhere
nonzero.  Multiplying these equalities by \eqref{i4'}, we get \eqref{i1'},
\eqref{i2'} and \eqref{i3'}, respectively. It is now a matter of routine to
prove that the independence of $\ce{(a|b)}_n$ and $\ce{(c|d)}_n$ follows from
\eqref{i1'},
\eqref{i2'} and \eqref{i3'} and \eqref{i4'}.
\end{proof}

The lemma allows us to characterize independence for present tense
conditionals.

\begin{theorem}\label{indep_present}
 For present tense $(a|b)$ and $(c|d)$ the following are
equivalent: 
\begin{itemize}
\item Stochastic processes $\ce{(a|b)}$ and $\ce{(c|d)}$ are independent.
\item The equalities \eqref{i1}--\eqref{i4} hold with $\pr_n$ replaced by
$\Pr$ in each term, where we again assume an equation to hold in case when
both sides are undefined.
\end{itemize}
\end{theorem}

\begin{proof} For present tense $(a|b)$ the probability $\pr_n((a|b))$ is
independent of $n,$ and is (of course) equal to $\Pr((a|b)).$ Now
Lemmas \ref{present-tense} and \ref{indep} give us the desired
equivalence.
\end{proof}

The full characterization of independence for general temporal
conditionals is not known at the moment. Most likely, if it at all
exists, it must be nonuniform, in the sense that the number of
equalities between probabilities depends in principle on the actual
$(\varphi|\psi)$ and $(\zeta|\xi).$

However, there is a quite general sufficient condition for
independence, which can be (nonuniformly) characterized by equalities
of asymptotic probability.

Call two conditionals $a,b$ {\em strongly independent\/} iff there
exist stochastically independent Markov chains $\X$ and $\Y$ and
projections $h,g$ such that $\ce{a}=h(\X)$ and $\ce{b}=g(\Y).$

\begin{theorem}\label{indep_strong} Strong independence of conditional
events from $\TT$ can be equivalently characterized by equations of asymptotic probability.
\end{theorem}

We begin with

\begin{lemma}\label{MN}
Let Markov chains $\X,\Y$ have $n$ and $m$ states, respectively. If
$\X$ and $\Y$ are independent until time $mn+1,$ they are fully
independent, i.e., if

\begin{multline}\label{mn}
\Pr\left(\hspace{-6pt}
\begin{array}{ccc}
X_1=w_1,&\dots,&X_t=w_t\\
Y_1=v_1,&\dots,&Y_t=v_t
\end{array}
\hspace{-6pt}\right)=\\
\Pr(X_1=w_1,\dots,X_t=w_t)\Pr(Y_1=v_1,\dots,Y_t=v_t)
\end{multline}

holds for all $t\leq mn+1$ and all sequences $w_1,\dots,w_t,$
$v_1,\dots,v_t$ of states of $\X$ and $\Y,$ respectively, then $\X$
and $\Y$ are independent and \eqref{mn} holds indeed for all $t.$
\end{lemma}
\begin{proof}
First of all, observe that $(\X,\Y)=(X_1,Y_1),(X_2,Y_2),\dots$ is a
Markov chain, as well. 

Suppose that \eqref{mn} fails and that the least $t$ for which it
fails is $t>mn+1$ (because for $t\leq mn+1$ \eqref{mn} holds by
assumption).  

The in-equality

\begin{multline}\label{in-eq}
\Pr\left(\hspace{-6pt}\begin{array}{ccc}
X_1=w_1,&\dots,&X_t=w_t\\
Y_1=v_1,&\dots,&Y_t=v_t
\end{array}\hspace{-6pt}\right)\neq\\
\Pr(X_1=w_1,\dots,X_t=w_t)\Pr(Y_1=v_1,\dots,Y_t=v_t)
\end{multline}

is by Markov property \eqref{MP} for $\X,$ $\Y$ and $(\X,\Y)$
equivalent to

\begin{multline*}
\Pr\left(\hspace{-6pt}
\begin{array}{ccc}
X_1=w_1,&\dots,&X_{t-1}=w_{t-1}\\
Y_1=v_1,&\dots,&Y_{t-1}=v_{t-1}
\end{array}
\hspace{-6pt}\right)
\Pr\left(\left.\hspace{-6pt}
\begin{array}{c}
X_{t}=w_{t}\\
Y_{t}=v_{t}
\end{array}
\right|
\begin{array}{c}
X_{t-1}=w_{t-1}\\
Y_{t-1}=v_{t-1}
\end{array}\hspace{-6pt}
\right)
\neq\\
\Pr(X_1=w_1,\dots,X_{t-1}=w_{t-1})\Pr(X_t=w_t|X_{t-1}=w_{t-1})\times\\
\Pr(Y_1=v_1,\dots,Y_{t-1}=v_{t-1})\Pr(Y_t=v_t|Y_{t-1}=v_{t-1}),
\end{multline*}

which in turn is equivalent to 

\begin{multline}\label{sprzecznosc}
\Pr\left.\left(\hspace{-6pt}\begin{array}{c}
X_{t}=w_{t}\\
Y_{t}=v_{t}\end{array}\right|
\begin{array}{c}
X_{t-1}=w_{t-1}\\
Y_{t-1}=v_{t-1}
\end{array}\hspace{-6pt}\right)
\neq\\
\Pr(X_t=w_t|X_{t-1}=w_{t-1})
\Pr(Y_t=v_t|Y_{t-1}=v_{t-1}),
\end{multline}

because $t$ is the least one for which in-equality holds, and so the
non-conditional probabilities in the previous in-equality cancel out.

Moreover, the canceling terms must be nonzero for the in-equality to
hold, which means $(w_{t-1},v_{t-1})$ is reachable with positive probability
from the initial state in $(\X,\Y).$ But therefore it must be
reachable with positive probability in at most $mn$ steps, because
there are exactly so many states in $(\X,\Y).$ So let $(x_i,y_i),\
i=1,\dots, s\leq mn$ be a sequence of states of $(\X,\Y)$ leading to
$(x_s,y_s)=(w_{t-1},v_{t-1})$ with positive probability. By assumption 

\begin{multline}
\Pr\left(\hspace{-6pt}\begin{array}{ccc}
X_1=x_1,&\dots,&X_s=x_s\\
Y_1=y_1,&\dots,&Y_s=v_s
\end{array}\hspace{-6pt}\right)=\\
\Pr(X_1=x_1,\dots,X_s=x_s)\Pr(Y_1=y_1,\dots,Y_s=v_s),
\end{multline}

because $s\leq mn.$ If we now multiply the above by
\eqref{sprzecznosc}, we get, by a calculation reverse to what we have
done above, an instance  of \eqref{in-eq} with $t\leq mn+1,$ a
contradiction.\end{proof}

\begin{lemma} 
For given Markov chains $\X$ and $\Y$ and for a fixed time $t,$ fixed
sequences $w_1,\dots,w_t$ and $v_1,\dots,v_t$ of states of $\X$ and
$\Y,$ respectively, the formula \eqref{mn} can be equivalently
characterized by equalities among asymptotic probabilities of certain
conditionals, derived from $\X$ and $\Y.$
\end{lemma}
\begin{proof} Let $\A$ and $\B$ be the deterministic finite automata,
underlying $\X$ and $\Y.$ For a state $w$ of $\A$ let $\A_w$ be the Moore
machine resulting from $\A$ by labeling the state $w$ with 1 and all the
remaining states with 0.  Since all $\A_w$'s are $\dwa$-valued, there exist
$\TL$ formulas $\alpha_w,$ which are true precisely when the last symbol of
the output of $\A_w$ is 1.  Similarly we define $\B_v$ and $\beta_v.$

Now \eqref{mn} is equivalent to

\begin{multline*}
\Pr(\PREV^{t}\TRUE\land\lnot\PREV^{t+1}\TRUE\land
\Land_{i=1}^{t}(\PREV^{t-i}(\alpha_{w_{i}}\land\beta_{v_{i}})))=
\\
\Pr(\PREV^{t}\TRUE\land\lnot\PREV^{t+1}\TRUE\land
\Land_{i=1}^{t}(\PREV^{t-i}(\alpha_{w_{i}})))\times\\
\Pr(\PREV^{t}\TRUE\land\lnot\PREV^{t+1}\TRUE\land
\Land_{i=1}^{t}(\PREV^{t-i}(\beta_{v_{i}}))).
\end{multline*}

Each of the $\TL$ formulas asserts that it has been once that there
was something $t-1$ steps ago, but there was nothing $t$ steps ago (so
we have been at time $t$ precisely), and we were in the prescribed
states of the Markov chain in question at times: $t$, one step before
that, \ldots, $t-1$ steps before that.

\end{proof}

What remains to be seen is that we can indeed choose some canonical
Markov chains $\X$ and $\Y$ to represent $a$ and $b,$ which are
independent whenever $a$ and $b$ are strongly independent.

Let us recall, that any conditional event in our model is a projection
of a Markov chain, derived from a Moore machine for the underlying
conditional object. Since for every Moore machine there exists the
minimal Moore machine computing the same function, in presence of
probabilities, we thus always have the minimal Markov chain underlying
any given conditional event.

\begin{lemma}\label{minimal}
Let $a$ and $b$ be strongly independent. Then the minimal Markov chains 
for $a$ and $b$ are independent.
\end{lemma}
\begin{proof} Let $\X$ and $\Y$ be two independent Markov chains, underlying
$a$ and $b.$ Applying the quotient construction to $\X$ and $\Y$ we
pass to the minimal Markov chains underlying $a$ and $b.$ The quotient
construction is deterministic, and therefore it does not break
independence (exactly like strong independence implies independence).
It follows that the minimal Markov chains are independent, too.
\end{proof}

\begin{proof}[Proof of Theorem \ref{indep_strong}] 

The conditional events $a$ and $b$ are strongly independent iff the
minimal Markov chains underlying them are independent, by Lemma
\ref{minimal}. The latter can be expressed equivalently by
$(mn+1)^{mn}$ conditions of the form \eqref{mn} for minimal chains of
$m$ and $n$ states, respectively, by Lemma \ref{MN}. Each of these
conditions in turn can be expressed equivalently by a single equality
of asymptotic probabilities of certain conditional objects.  This
means that the strong independence of $a$ and $b$ can be equivalently
characterized by a set of equalities among asymptotic probabilities of
conditionals, which can be syntactically determined from $a$ and $b$
and do not depend on the probability space structure.
\end{proof}

Of course, for conditionals which are themselves Markov chains for any
probability assignment, strong independence is the same as independence.
Therefore we have

\begin{corollary} For conditionals which are themselves Markov chains
for any probability assignment, independence can be characterized by
equalities of asymptotic probabilities.
\qed\end{corollary}

The conditionals to which this applies can be recognized by the
property that their minimal Moore machine has at most one state
labeled by each element of $\trzy$ (and thus at most three states
altogether). Present tense conditionals are of this kind, and thus we
have an alternative proof of Theorem \ref{indep_present}, which much
less elegant set of equalities, however. But present tense
conditionals do not exhaust all conditionals, which are Markov
chains. An example is the conditional $(a|\PBOX ((\PREV a\to
a^\C)\land(\PREV a^\C\to a)\land(\lnot\PREV \TRUE\to a ))),$ analyzed
in \cite{TT}. Its minimal Moore machine is depicted below.

\begin{figure}[h!]

\[\UseTips
\xymatrix @C=20mm
{
&*+++[o][F]{1}
\ar@/^/[dr]^{\textstyle{a}}
\ar@<-.4em>[dd]_{\textstyle{a^\C}}
\\
&&*+++[o][F]{\bot}
\ar@(ur,dr)[]
\\
\ar[r]
&*+++[o][F]{0}
\ar@/_/[ur]_{\textstyle{a^\C}}
\ar@<-.4em>[uu]_{\textstyle{a}}
}
\]

\caption[Non-simple conditional with 3-states]{Moore machine of
$(a|\PBOX ((\PREV a\to a^\C)\land(\PREV a^\C\to a)\land(\lnot\PREV
\TRUE\to a ))).$}\label{f_indep}
\end{figure}

Therefore Theorem \ref{indep_strong} is indeed stronger than Theorem
\ref{indep_present}.

Finally, we consider the question of $\PS$ \cea, for which one might
want a characterization of independence in terms of asymptotic
probability. Here we give a negative answer.

\begin{theorem}\label{(a|0)} There exist two conditional expressions
$e_1$ and $e_2$ and a probability space such that the embeddings
$\ttt{e_1}$ and $\ttt{e_2}$ are independent, while their sparse reverse
counterparts $\ttrs{e}$ and $\ttrs{e_2}$ are not independent.
\end{theorem}
\begin{proof} Take $e_1=e_2=(0|a)$ and any probability space with
$0<\Pr(a)<1.$ Then $\ce{\ttt{e_1}}$ and $\ce{\ttt{e_2}}$ are constant
processes, equal to $0,$ so they are (trivially) independent. However,
already

\begin{align*}\Pr(\begin{array}{c}\ce{\ttrs{e_1}})(w)=0\bot
\\\ce{\ttrs{e_2}})(w)=0\bot\end{array})&=
\Pr(\ce{\ttrs{e_1}})(w)=0\bot)\\
&>(\Pr((\ce{\ttrs{e_1}})(w)=0\bot))^2\\
&=\Pr(\ce{\ttrs{e_1}})(w)=0\bot)\cdot
\Pr(\ce{\ttrs{e_2}})(w)=0\bot),
\end{align*}

where the inequality holds because
$\Pr(\ce{\ttrs{e_1}})(w)=0\bot)=\Pr(a)(1-\Pr(a))\neq 0,1.$\end{proof}

\begin{corollary} \label{PS-indep} There is no characterization of
independence in $\PS$ \cea\ in terms of equalities of asymptotic
probability.
\end{corollary}
\begin{proof} Because both $\ttt{\cdot}$ and $\ttrs{\cdot}$ preserve
all asymptotic probabilities of conditionals, both of them satisfy
precisely the same equalities of asymptotic probabilities. So if there
were a characterization of independence in terms of equalities of such
probabilities, the interpretations of the two conditionals $(a|0)$ and
$(a|0)$ above would have to be either independent in both cases, or
dependent in both cases, while they are not, a contradiction.
\end{proof}

The consequence is that in the \cea{} formalism is not expressive
enough to define independence of conditionals by means of equalities
of asymptotic probabilities. Note however, that such a representation
is certainly possible by means of equalities of probabilities and
equalities of the algebraic structure. Indeed, $\PS$ \cea{} is boolean
algebra with respect to its connectives $\land,\lor,\llnot$ (as it is
easily visible from its syntactic representation within $\TT$), and
the equalities it satisfies enforce, that $\Pr_o$ is an ordinary
probability measure. Therefore independence is equivalent to the
standard equality $\Pr_o(\ps{e_1\land
e_2})=\Pr_o(\ps{e_1})\Pr_o(\ps{e_2}).$ What we have constructed are
two non-boolean subsystems of $\TT$, in which all the (asymptotic)
probability assignments agree with those of $\Pr_o$, and yet no set of
equalities of probabilities can characterize the true probabilistic
independence in both of them simultaneously.

\subsection{Algorithms}\label{algorytmy}

\paragraph{Polynomial algorithm for PS \cea.} Let us see that our
approach provides a nontrivial improvements to the algorithmic status of
existing \cea's. We will demonstrate this by calculating the
probabilities of conditional expressions, according to $\PS$ \cea, in
time polynomial in their size and exponential in the number of
variables. (Note that the number of arguments for computation of the
probability of an $n$-ary conditional is $2^n,$ so the above indeed
means computation polynomial in the size of the input.) In \cite{g94} it
is stated that the computation of the $\PS$-probability of a conjunction
of $n$ conditionals $(a_i| b_i),$ according to the method used in that
paper, requires adding $\sum_{m=1}^n m!\cdot S_0(m,n)\cdot (2^{m+1}-2)$
terms, each being a nonconditional probability of a conjunction of
certain events $a_i$ and $b_i.$ The number of summands, where $S_0(m,n)$
are Stirling's number of the second kind, is of order $2^{n\log n}.$ It
is substantially more than about $c2^{n}$ one obtains for the present
tense \cea's $\SAC$ and $\GNW,$ and has been stressed in
\cite[p.~499]{kniga} and in \cite{recenzja}, since it strongly affects
the usefulness of $\PS$ as a tool for applications. Using our approach we
have instantly an algorithm to calculate the same probability in
$2^{O(n)}$ steps. As a matter of fact, this applies to {\em any\/}
conditional expression with $n$ arguments $(a_i| b_i),\ i=1\dots,n,$ as
long as its length does not exceed $2^{O(n)}.$ All the complexity bounds
given here assume unit cost of basic arithmetical operations: addition,
multiplication, subtraction and division.

\begin{theorem} There is an algorithm, computing the $\PS$\/ probability
of an $n$-ary conditional expression of length $m$ in time polynomial in
$\max(m,2^n).$
\end{theorem}
\begin{proof} The first step of the algorithm on input expression $e$ is
to construct the minimal Moore machine, computing $\ttt{e}.$ 

\begin{lemma} The minimal Moore machine of $\ttt{e}$ for $n$-ary
conditional expression $e$ has at most $3^n$ states.
\end{lemma}
\begin{proof} The minimal Moore of $\first(a|b)$ has 3 states (see
\eqref{first} and Figure \ref{f4}).

By the definition of the $\ttt{\cdot}$ embedding (Section \ref{FE}), a
Moore machine $\A=(Q,\Omega,\delta,h,q_0)$ of $\ttt{e}$ can be
constructed as follows:

The set $Q$ of states of $\A$ is the product $Q_1\times\dots\times Q_n$
of state sets of Moore machines $\A_i=(Q_i,\Omega,\delta_i,h_i,q_{0i})$
of all expressions $\first(a_i|b_i)$ occurring in $\ttt{e}.$ The transition
function of $\A$ is defined coordinate-wise, i.e.,

\[\delta(\langle q_1,\dots,q_n\rangle,\omega)=\langle
\delta_1(q_1,\omega),\dots,\delta_n(q_n,\omega)\rangle,\]

the initial state is $q_0=
\langle q_{01},\dots,q_{0n}\rangle,$ and, crucially,

\[h(\langle
q_{1},\dots,q_{n}\rangle)=\hat{c}(h_1(q_1),\dots,h_n(q_n)),\]

where $\hat{c}(h_1(q_1),\dots,h_n(q_n))$ is the classical logic
evaluation of the expression $c$ on arguments
$h_1(q_1),\dots,h_n(q_n)\in\dwa.$

This product construction is well-known for automata theory, and it is
immediate that it does the work.
\end{proof}

So it is quite easy to construct, given $e\in\LL,$ the Moore machine of
$\ttt{e}.$ Now we have to turn this Moore machine into a Markov chain.
Assuming that all the probabilities of atomic events from $\Omega$ are
given, we simply replace multiple transitions between the same states
represented by the sum of their probabilities---a single number.

Furthermore, the Markov chain we obtain is absorbing, i.e., it has
one-element ergodic classes. It can be proven by a straightforward
induction on $n$ --- the number of three element Moore machines we
product. It follows \cite[Chapter III]{KS} that we can use the following
method to compute the limiting probability that the chain finally
arrives at a state labeled by $1$.

Clearly, in this situation we can collapse all absorbing states labeled
$1$ into a single such state.

Denote by $P$ the matrix $(p(i,j))$ of transition probabilities, by $Q$
the submatrix of rows and columns corresponding to transient states, and
by $R$ the submatrix of rows corresponding to transient states and
columns corresponding to absorbing states. Let $Id$ be a diagonal matrix
with $1$'s on the diagonal and $0$'s elsewhere. Let $B=(Id-Q)^{-1}R.$
Then the probability we are looking for is the entry in $B$ in the row
corresponding to the initial state in in the column corresponding to the
(only) absorbing state labeled by $1$ in the Markov chain. Since all
the calculations on matrices necessary to compute $B$ are doable in time
polynomial in the size of the matrices, the total computation time is
$(2^{n})^{O(1)}=2^{O(n)},$ as desired.
\end{proof}

\section{Summary}

We have discussed the temporal calculus of conditional objects and
conditional events $\TT$ as a formalism alternative to conditional event
algebras.

We have shown that all the major conditional event algebras, including
those of Schay-Adams-Calabrese, Goodman-Nguyen-Walker and the product
space \cea, embed isomorphically in $\TT.$

Moreover, $\TT$ is superior to those formalisms in several ways:

\begin{itemize}
\item It provides natural, probabilistic semantics of conditionals,
allowing one to construct experiments to evaluate all their interesting
probabilistic parameters, unlike \cea's, which generally are not
probability spaces, and which do not require certain probabilistic
parameters to be defined at all.
\item The construction of $\TT$ is functorial, in the sense, that the
underlying probabilistic space of nonconditional events determines the
space of temporal conditional events uniquely, while \cea's generally
are not unique.
\item The formalism of $\TT$ allows one to define and analyze
independence of conditional events, which is difficult or impossible in
\cea's.
\item $\TT$ offers better algorithms for calculation of probabilities,
than those known previously for \cea's.
\end{itemize}


\bibliographystyle{alpha} \bibliography{bib}


\end{document}